\documentclass[11pt,letterpaper]{article}
\usepackage[margin=1in]{geometry}
\usepackage[T1]{fontenc}
\usepackage{times}
\usepackage[numbers,sort&compress]{natbib}
\usepackage{graphicx}
\usepackage{booktabs}
\usepackage{multirow}
\usepackage{amsmath,amssymb}
\usepackage[dvipsnames]{xcolor}
\usepackage{microtype}
\usepackage{xspace}
\usepackage{placeins}
\usepackage[format=plain,labelformat=simple,labelsep=period,font=small]{caption}
\usepackage[hyphens]{url}
\definecolor{linkblue}{rgb}{0.21,0.49,0.74}
\usepackage[breaklinks,colorlinks,allcolors=linkblue]{hyperref}
\hypersetup{pdftitle={Seeing Like a Shopper: Validating Vision-Language Models as Synthetic Consumers},pdfauthor={Woongki Kim, Seungkook Baek, Minhyeok Lee}}
\usepackage[capitalize]{cleveref}
\crefname{section}{Sec.}{Secs.}
\Crefname{section}{Section}{Sections}
\crefname{table}{Tab.}{Tabs.}
\Crefname{table}{Table}{Tables}
\crefname{appendix}{App.}{Apps.}
\Crefname{appendix}{Appendix}{Appendices}

\newcommand{\bench}{ShelfSight\xspace}

\title{\bfseries Seeing Like a Shopper: Validating Vision-Language Models\\ as Synthetic Consumers}
\author{Woongki Kim\textsuperscript{1}\quad Seungkook Baek\textsuperscript{1}\quad Minhyeok Lee\textsuperscript{2,*}\\[6pt]
{\normalsize \textsuperscript{1}Intellicia Inc.}\\[1pt]
{\normalsize \textsuperscript{2}School of Electrical and Electronics Engineering, Chung-Ang University, Seoul, Republic of Korea}\\[3pt]
{\normalsize \texttt{wk.kim@intellicia.co}, \texttt{sk.baek@intellicia.co}, \texttt{mlee@cau.ac.kr}}\\[1pt]
{\normalsize \textsuperscript{*}Corresponding author}}
\date{}

\begin{document}
\maketitle

\begin{abstract}
Language models are widely used as synthetic respondents in surveys, behavioral experiments, and market research. Most of these respondents read text, whereas shoppers in a store choose by looking at shelves, which makes vision-language models (VLMs) natural synthetic shoppers. We test whether VLM shoppers follow real consumers at two levels. At the purchase level, VLM shoppers built from household records plan their trips from a store directory and select products from shelf images, and we compare their purchases with six public datasets from three countries. The synthetic baskets hit held-out purchases as often as the households' own past trips, and synthetic group differences correlate with real ones twice as strongly as those of a trained regression. Without fitting, VLM shoppers reproduce alcohol purchases by gender and age in Russia and trip sizes by store distance in Taiwan, and respond to situations, prices, and displays within measured ranges. At the level of a single shelf choice, a model's answer confounds the product with its position and letter label. We therefore introduce \bench, shelf renderings that vary products, positions, labels, viewpoints, and prices independently, together with a factor model that separates product preferences from the effects of label and position, which we call answer effects. On \bench, hosted VLMs read the shelf, follow stated needs, keep their product preferences across views, and respond to price as budget shoppers. These answer effects, a preference for the letter A and for positions in the image, are estimated once per image type and removed. VLMs are thus good synthetic shoppers whose corrected shelf choices reflect product preferences.
\end{abstract}

\begin{figure*}[t]
\centering
\includegraphics[width=\linewidth]{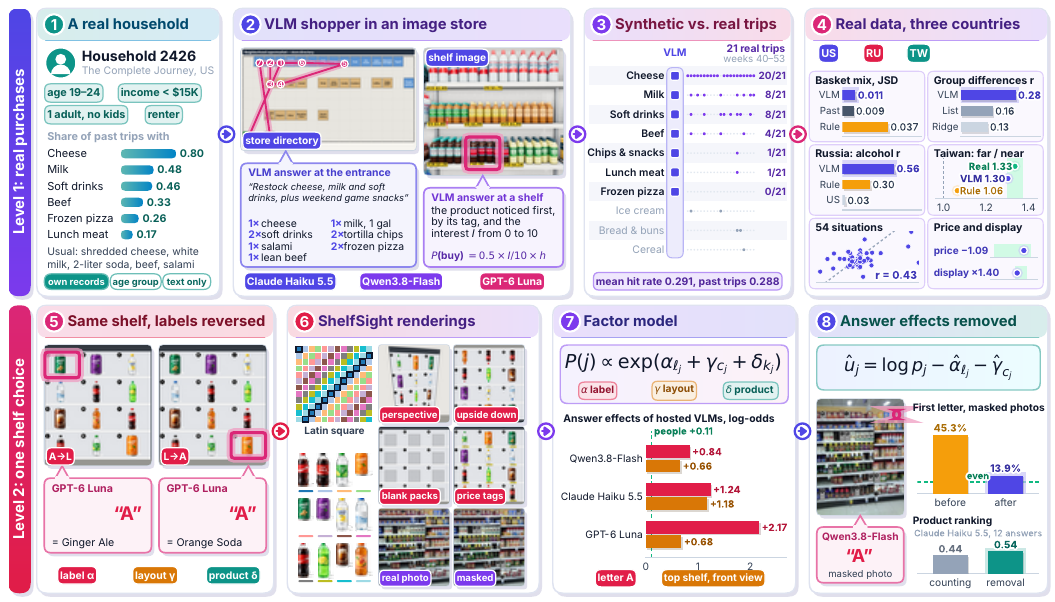}
\caption{Overview with a worked example. Purchase level: (1) a held-out household of The Complete Journey with the purchase rates of its records; (2) the answer of a VLM shopper at the store directory with its planned stops, and a shelf image with numbered tags (illustrated with product photos; the store packs carry printed names); (3) categories of its synthetic basket against its real test-period trips, with hit rates averaged over all households; (4) comparisons with real purchases from three countries (JSD: Jensen--Shannon divergence), with measured ranges in green. Shelf level: (5) answers of GPT-6 Luna on one ShelfSight shelf with labels in reading and reverse order; (6) ShelfSight renderings; (7) the factor model and the answer effects of the hosted models; (8) removal of the answer effects on masked real photos and on the product ranking from 12 answers per shelf of Claude Haiku 5.5.}
\label{fig:pipeline}
\end{figure*}

\section{Introduction}
\label{sec:intro}

Language models now stand in for human respondents in surveys, behavioral experiments, and market research \cite{argyle2023out,horton2023large,park2024generative,ashokkumar2026large}. In marketing, the models estimate willingness to pay, preference shares, and brand perceptions \cite{brand2023using,goli2024frontiers,li2024frontiers,sarstedt2024using} and simulate online shoppers \cite{lu2026can,wang2026opera,mansour2025paars} and users of recommender systems \cite{wang2025user,zhang2024generative}. Thousands of synthetic consumers can thus be asked before anyone is recruited.

Much of consumer choice, however, is made in front of a shelf. Shoppers scan packs, price tags, and promotion signs and decide many purchases in the store \cite{bell2011point}, where shelf position changes what shoppers notice and choose \cite{chandon2009does}. Many retail questions are visual, such as whether a new pack stands out, a layout lifts a category, or a sign moves sales. Text respondents cannot answer such questions, whereas a vision-language model (VLM) sees what a shopper sees, and platforms such as TwinStore \cite{twinstore2026} fill rendered stores with VLM shoppers. If VLM shoppers follow real consumers, packs, layouts, and prices can be pretested.

Two gaps keep VLM shoppers from this use. Synthetic respondents have been validated against surveys and controlled experiments and rarely against store purchase records \cite{gui2023challenge,sarstedt2024using}, and whether VLM shoppers buy like real households remains unknown. A single shelf choice is also ambiguous. Because a VLM names a product by a label placed on the image \cite{yang2023setofmark}, a choice of the top-left product labeled A can express a preference for the product, for the top of the image, or for the letter A. Language models favor particular option labels \cite{zheng2024large,pezeshkpour2024large} and VLMs favor particular image positions \cite{tian2025identifying,chaudhary2025investigating}, yet studies of synthetic respondents compare only final answers with people \cite{argyle2023out,brand2023using,goli2024frontiers}. A synthetic pack test that reads these answers at face value credits a new pack with the advantage of its shelf slot and its letter.

We close both gaps by testing VLM shoppers at two levels (\cref{fig:pipeline}). At the level of purchases, VLM shoppers built from the records or descriptions of real households and customers plan their trips from a store directory and select products from shelf images, and we compare their purchases with six public datasets from three countries, against baselines that use the same information without a language model. The synthetic baskets hit held-out purchases as often as the households' own past trips, and VLM shoppers given only age-group purchase rates reproduce demographic differences more strongly than a regression trained on 383 households. Without fitting to Russian or Taiwanese data, the shoppers recover consumption differences by gender and by distance to the store, and follow the situational, price, and display effects of real stores.

At the level of a single shelf choice, we introduce \bench, shelf renderings in which a Latin square moves the products through every position, the labels follow reading, reverse, or random order, and further renderings change the camera, packs, prices, or answer format one at a time. A conditional-logit factor model divides each choice into a product effect and two answer effects, a label effect and a layout effect. Hosted VLMs read the shelf, choose the product that meets a stated need, keep their product preferences across views and wording, and respond to price as budget shoppers. The answer effects of these models are a preference for the letter A and a preference for the top of the image beyond that of people, which explains the shelf-height preference of VLM shoppers in the store. Because both effects are shared across categories, one estimate per image type removes the answer effects without training, from one letter distribution or from repeated answers. On held-out categories the removal ranks products better than existing calibration methods and better than counting the same answers.

Our contributions are threefold.
\begin{itemize}
\item \textbf{A VLM synthetic shopper} that sees the store as images, validated on real purchases from three countries, whose group differences correlate with real ones twice as strongly as those of a trained regression.
\item \textbf{\bench}, controlled shelf renderings with a factor model that separates and removes answer effects without training, lowering the first-letter share of Qwen3.8-Flash on masked real photos from 45.3\% to 13.9\% (even choice 9.0\%).
\item \textbf{Evidence} that VLMs follow real consumers, from household baskets and market differences to single shelf choices, and a measurement of the answer effects of VLMs, which raise the odds of the letter A up to 8.8 times and favor the top of the image several times more than people do.
\end{itemize}

\section{Related Work}
\label{sec:related}

\paragraph{Language models as synthetic respondents.}
A language model conditioned on the profile of a respondent reproduces the answer distributions of survey subgroups \cite{argyle2023out}, replicates economic and psychological experiments \cite{horton2023large,aher2023using,xie2024can}, predicts opinions missing from surveys \cite{kim2023aiaugmented}, and forecasts the results of social science experiments \cite{ashokkumar2026large}. Generative agents simulate daily life \cite{park2023generative} and the survey answers of 1,000 people \cite{park2024generative}, and agent-based simulation covers many social domains \cite{gao2024large}. Studies of the limits report that model opinions lean toward some groups \cite{santurkar2023whose}, that synthetic answers vary less than human ones and change with the prompt \cite{bisbee2024synthetic,dillion2023can}, and that changing one attribute in a prompt can shift other inferred attributes \cite{gui2023challenge}. We compare VLM shoppers with purchase records instead of surveys and separate product preferences from answer effects.

\paragraph{Synthetic consumers and simulated stores.}
Marketing studies use language models to estimate willingness to pay and preference shares \cite{brand2023using,goli2024frontiers} and brand perceptions \cite{li2024frontiers}, with guidelines for consumer research \cite{sarstedt2024using}. Agents simulate users of recommender systems \cite{wang2025user,zhang2024generative}, online shoppers compared with real sessions \cite{lu2026can,wang2026opera}, and retail shoppers with personas from purchase histories \cite{mansour2025paars}. The purchase of an AI shopping agent depends on where a product appears in a listing \cite{allouah2025what}. Web agents act on screenshots \cite{yao2022webshop,koh2024visualwebarena,zheng2024gpt4v}, usually through set-of-mark labels \cite{yang2023setofmark}. Store digital twins \cite{tao2019digital,hornik2026designing} build on path data \cite{hui2009path} and on findings that shelf height changes attention \cite{chandon2009does}, that complementary categories placed together sell more \cite{bezawada2009cross}, and that many purchases are decided in the store \cite{bell2011point}. VLMs also recognize shelf products \cite{goldman2019precise,tur2024exploring,pietrini2024shelf}. Our shoppers instead select products from store images and are compared with purchase records.

\paragraph{Position and label bias in model choices.}
Language models prefer particular option labels in multiple-choice questions \cite{zheng2024large,pezeshkpour2024large}, favor the first answer when judging pairs \cite{wang2024large}, and use the ends of a long context more than the middle \cite{liu2024lost}. Content-free calibration \cite{zhao2021calibrate}, batch calibration \cite{zhou2024batch}, and PriDe, which estimates the label prior over option permutations \cite{zheng2024large}, reduce such biases for text. VLMs favor some image positions \cite{tian2025identifying,chaudhary2025investigating} and respond to marks placed on an image \cite{shtedritski2023what}. On a labeled shelf, label order and image position are confounded by construction, and \bench separates the two from each other and from product content.

\paragraph{Visual attention on shelves.}
Shoppers fixate the center of a display and the middle shelves more often, additional facings raise attention and choice, and attention predicts choice \cite{chandon2009does,atalay2012shining,gidlof2017looking}. Models of where people look range from spectral-residual saliency \cite{hou2007saliency} to learned fixation predictors \cite{judd2009learning,kummerer2022deepgaze,droste2020unified}, and the central fixation bias \cite{tatler2007central} is a strong baseline for scenes viewed head-on. We use these measurements as human references and to fit an attention prior.

\section{VLM Synthetic Shoppers}
\label{sec:shopper}

A synthetic shopper corresponds to one real household or customer and makes one shopping trip at a time in a store simulated by TwinStore \cite{twinstore2026}. The simulator handles walking and dwell time, and a VLM decides what to buy (\cref{fig:pipeline}).

\paragraph{Personas.}
A persona gives the demographics and home location of a household (age group, income, children, household size, marital status) and summarizes the household's records before the test period, including the purchase rate and the usual product of each category. Before each trip, categories that the household needs are sampled from these rates and listed for the model as suggestions. The model decides how each trip departs from the suggestions, adding knowledge beyond the records. In the \emph{own-records} condition the persona carries the household's own rates and usual products, as a retailer's loyalty data would. In the \emph{age-group} condition the persona carries only the average rates of the household's age group, and income, children, size, and marital status reach the model only in text. This condition isolates what the model infers from a description beyond the purchase rates. In the unseen markets, the persona holds only a customer description, and the model sets the trip size.

\paragraph{A store of images.}
The store holds the 30 best-selling categories of The Complete Journey \cite{dunnhumby_completejourney}, a US panel of 801 households with demographics, and eight products in each category at representative prices. The store directory shows the shelves with category names and marks promoted shelves. Each shelf image shows the products with numbered price tags and as packs with printed names, which keep every product identifiable. With one store for every country, differences between groups come from the shoppers.

\paragraph{Decisions.}
At the entrance the model receives the persona, the time, the situation, a budget, the directory image, and the product list, and states the purpose of the trip and the categories, products, and quantities to buy. Along the route the shopper views nearby shelves. For each shelf image the model names the product that the shopper notices first and rates the shopper's interest in that product from 0 to 10, and the product is bought with probability $0.5 \times \text{interest}/10 \times h$, where $h$ falls from 1.00 at eye level to 0.55 on the bottom shelf. The model thus decides which product a shopper wants and how strongly, and a rule shared by all shoppers converts this interest into purchases. Situations such as ``Sunday with a big game'' are given to the model as text, and measured situational responses can also weight the sampled categories.

\paragraph{Models.}
Shoppers are assigned at random to three hosted VLMs, Claude Haiku 5.5, Qwen3.8-Flash, and GPT-6 Luna, and the results describe this mixed population.

\section{Validation on Real Purchases}
\label{sec:real}

\begin{table}[t]
\centering
\footnotesize
\setlength{\tabcolsep}{4pt}
\resizebox{\ifdim\width>\linewidth\linewidth\else\width\fi}{!}{%
\begin{tabular}{@{}llll@{}}
\toprule
Dataset & & Records & Use \\
\midrule
The Complete Journey \cite{dunnhumby_completejourney} & US & 801 households & baskets, groups \\
X5 RetailHero \cite{x5retailhero} & RU & 8.0M receipts & unseen market \\
Ta-Feng \cite{tafeng} & TW & 119,578 receipts & unseen market \\
Instacart \cite{instacart2017} & US & 3.4M online orders & next order \\
M5 \cite{makridakis2022m5} & US & 6.8M item-weeks & price \\
Breakfast at the Frat \cite{dunnhumby_frat} & US & 77 stores, 156 weeks & promotion \\
ATC \cite{brscic2013person} & JP & 13,159 tracks & walking speed \\
Eye tracking \cite{hummel2021visual} & DE & 16 shoppers & dwell time \\
\bottomrule
\end{tabular}}
\caption{Public datasets from five countries and their use.}
\label{tab:datasets}
\end{table}

\paragraph{Data and protocol.}
\Cref{tab:datasets} lists the eight public datasets. Six datasets test the purchases of VLM shoppers, and ATC and the eye-tracking study calibrate the walking speed and dwell time of the simulator. In The Complete Journey, the simulator is calibrated on households outside the test panels, and two held-out panels of 150 households each receive personas built from weeks 1 to 39 and are compared with the real purchases of weeks 40 to 53. For X5 and Ta-Feng we sample 30 customers from each of 12 age-and-gender or age-and-residence groups and keep the store and the simulator of The Complete Journey unchanged. We judge each market by one criterion. In X5 the correlation with the real group effects must be positive at permutation $p<0.05$, and in Ta-Feng the far-to-near ratio must have an interval above one and lie closer to the real ratio than the transferred value.

\paragraph{Baselines.}
\emph{Past trips} replays a real earlier trip of the same household. A \emph{rate list} fills the same number of items from the same purchase rates without a language model. \emph{Learned} models predict category rates from seven demographic attributes by ridge regression or $k$-nearest neighbors (kNN), trained on the 383 households outside all panels. A \emph{transferred} model applies the age gradient of The Complete Journey to the other markets. A \emph{rule-based} shopper reads the same persona and executes a fixed category list with set probabilities and one price elasticity. Because each baseline receives the information of the VLM shopper in the same condition, the gap measures what the language model adds.

\begin{figure*}[t]
\centering
\includegraphics[width=\linewidth]{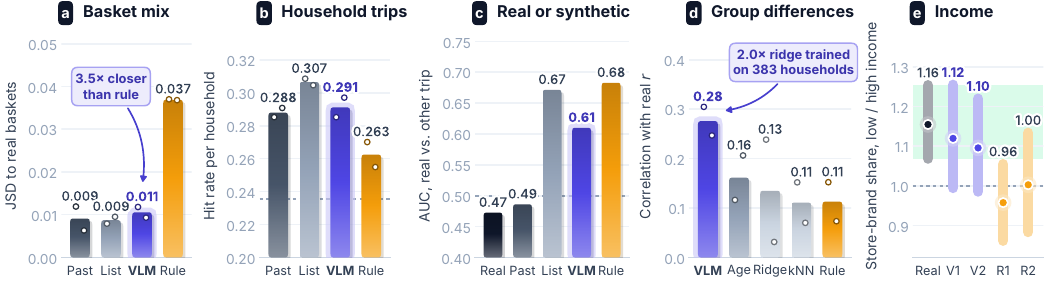}
\caption{Synthetic purchases against held-out real purchases of The Complete Journey. (a) Jensen--Shannon divergence between the category mixes of synthetic and real baskets. (b) Share of a household's real test-period trips that contain each category of its synthetic basket; the dashed line marks baskets sampled by store-wide purchase rates. (c) AUC of a classifier that distinguishes a real trip of a household from a trip of the same household made by each shopper or baseline, using category, size, and time features. Panels (a--c) use each household's own records. (d) Correlation between synthetic and real deviations of demographic groups from the mean category shares, given age-group average purchase rates. (e) Store-brand share of low-income over high-income households, with 95\% intervals, for VLM (V) and rule-based (R) shoppers in the two held-out panels (1, 2); the green band is the real interval. Past, List, and Rule mark past trips, the rate list, and the rule-based shopper, and Age, Ridge, and kNN mark the age-group rate list and the learned models. Bars average two panels of 150 households (dots).}
\label{fig:purchases}
\end{figure*}

\paragraph{Households.}
With the households' own records, VLM shoppers fill baskets whose category mix is as close to the real one as past trips and the rate list, and several times closer than that of the rule-based shopper (\cref{fig:purchases}(a)). The rate list replays the records by construction, whereas the VLM receives the rates only as suggestions and keeps this information while deciding each trip. VLM baskets hit the household's real trip categories as often as a past trip does (0.291 vs.\ 0.288, \cref{fig:purchases}(b)). Planning each trip around a stated purpose, VLM shoppers make trips that a classifier tells from real ones less well than rate-list or rule-based trips (\cref{fig:purchases}(c)), also with product-level features. In Instacart orders, VLM shoppers hit the aisles of a user's next order nearly as often as the user's own history and more often than popular aisles or the rule-based shopper, with each hosted model at the same level. In the age-group condition the model sees no household records, yet the categories added beyond the suggested list are bought in the real test period more often than categories with the same base rate (0.166 vs.\ 0.136) in both panels, evidence of knowledge beyond rates.

\paragraph{Differences between consumers.}
With only age-group rates, VLM shoppers reproduce how purchases differ by age, income, children, household size, and marital status (\cref{fig:purchases}(d)). The synthetic differences correlate with the real ones at 0.30 and 0.25 in the two panels, twice the 0.13 of ridge regression trained on 383 households and above the age-group rate list. The regression receives the demographic attributes as indicators and must learn each effect from data, whereas the VLM receives the same attributes as a description such as ``a low-income household with children'' and draws on prior knowledge of such households. At the product level, low-income households buy 1.16 times the store-brand share of high-income ones, and VLM shoppers given income only in text reproduce this ratio in both panels, whereas the rule-based shopper shows none (\cref{fig:purchases}(e)).

\begin{figure*}[t]
\centering
\includegraphics[width=\linewidth]{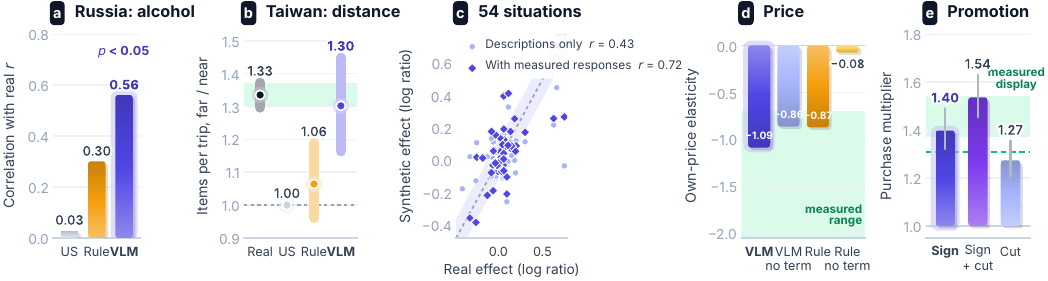}
\caption{Unseen markets, situations, price, and promotion. (a) Correlation with real data across 12 age and gender groups of X5 in the share of trips with alcohol; US marks the model transferred from the US data. (b) Items per trip of customers living far from the store over those living near the store in Ta-Feng, with 95\% intervals; the green band is the real interval. (c) Log ratios of purchase rates under six situations, one point per situation and category. (d) Own-price elasticity when half of the products change price by 20\%; ``no term'' marks shoppers without the fixed elasticity term, and the green band spans measured elasticities. (e) Purchase multiplier of promoted products with 95\% intervals; the green band spans measured display effects, and the green line marks a measured price cut without a display.}
\label{fig:transfer}
\end{figure*}

\paragraph{Unseen markets.}
In Russian X5 data, VLM shoppers given only age and gender reproduce the higher alcohol purchases of men and middle-aged customers across groups ($r=0.56$), meeting the criterion, while the transferred model without a gender term shows no such pattern (\cref{fig:transfer}(a)). In Taiwanese Ta-Feng data, customers who live far from the store buy 1.33 times as many items per trip as those living near the store, and VLM shoppers given only age and residence reproduce the ratio at 1.30, while the rule-based shopper barely responds (\cref{fig:transfer}(b)). Because the US statistics contain no gender and no distance to the store, both effects come from the model's knowledge of these markets.

\paragraph{Situations, price, and promotion.}
Given only descriptions such as ``Sunday with a big game'', VLM shoppers change their baskets in line with 54 real situational effects ($r=0.43$), while the rule-based shopper, which cannot read the descriptions, does not respond (\cref{fig:transfer}(c)). The response reflects the model's knowledge of such days. When half of the products change price, the own-price elasticity of VLM shoppers is $-1.09$, between $-0.70$ for Walmart items \cite{makridakis2022m5} and the average of $-2.62$ across published studies \cite{bijmolt2005new}. Without the fixed elasticity term of the simulator, VLM shoppers keep most of this response, whereas the rule-based shopper, which shares the budget and the store, loses nearly all of its response (\cref{fig:transfer}(d)). The price response is thus a decision of the model and not a product of the budget. A ``SPECIAL'' sign raises the purchases of a product 1.40 times, within the display effects of 1.37 to 1.54 measured at two US retailers \cite{dunnhumby_completejourney,dunnhumby_frat}, and a price cut without a sign raises purchases about as much as real temporary price cuts \cite{dunnhumby_frat} (\cref{fig:transfer}(e)).

\begin{figure}[t]
\centering
\includegraphics[width=0.62\linewidth]{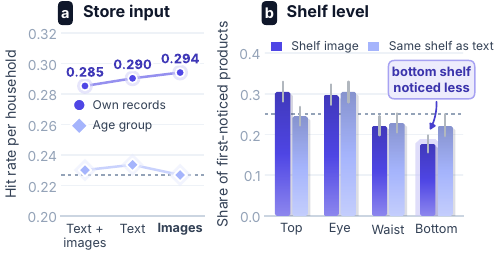}
\caption{The store as images. (a) Household hit rate in the first held-out panel when the store is given as a text list with images, as text only, or as the directory and shelf images only; the dashed line marks baskets sampled by store-wide rates. (b) Shelf level of the product that the shopper notices first when the products of each category rotate through the four levels, with 95\% intervals; the dashed line marks no preference.}
\label{fig:images}
\end{figure}

\paragraph{The store as images.}
Given only the directory and shelf images, VLM shoppers match the household hit rate of the text representations (\cref{fig:images}(a)). A synthetic shopper can therefore take the store as images alone, the form in which a new pack or layout is designed. When the products of each category rotate through the four shelf levels without the height weight $h$, VLM shoppers that receive the shelf image notice products on the bottom shelf less often than an even preference would, and this avoidance weakens when the same shelf is given as text (\cref{fig:images}(b)). The avoidance thus arises from the image, in the direction of human attention \cite{chandon2009does}, and \cref{sec:shelfsight} measures the size of this avoidance and removes the part beyond that of people.

\section{Separating Product Preferences from Answer Effects}
\label{sec:shelfsight}

\bench isolates the shelf step of a VLM shopper as a controlled choice from a labeled shelf image, in which letter labels take the place of the numbered tags of the store. We state five requirements for a good synthetic shopper at this step, each testable on images. The shopper must read the shelf, finding a product and its price (R1), and choose the product that meets a stated need (R2). Preferences must depend on the products and remain unchanged when the camera, the labels, or the wording change (R3). The shopper must respond to what the products look like and cost, as a shopper with the stated profile would (R4). Any dependence of the choices on labels or positions beyond that of people must be measurable and removable (R5).

\subsection{The \bench Benchmark}
\label{sec:bench}
\begin{figure*}[t]
\centering
\includegraphics[width=\linewidth]{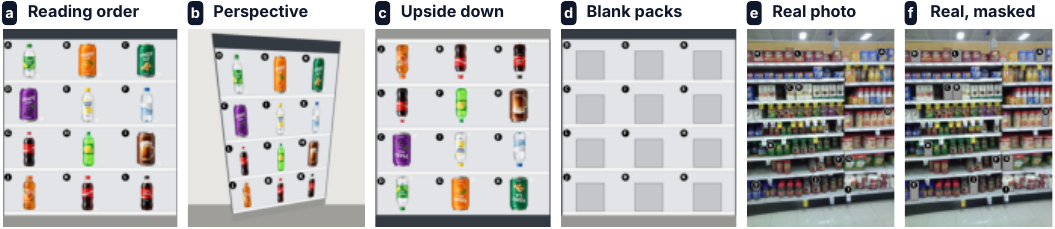}
\caption{Controlled renderings of one shelf of 12 products. (a) Product photos with letter labels in reading order. (b) The same products under a perspective camera at a shopper's eye height, with labels in random order. (c) The front view turned upside down with upright labels. (d) Identical blank packs that carry no product information. (e) A real shelf photo with twelve labeled products, one per region of a four-by-three grid. (f) The same photo with the labeled products painted gray.}
\label{fig:stimuli}
\end{figure*}
\begin{figure*}[!t]
\centering
\includegraphics[width=0.92\linewidth]{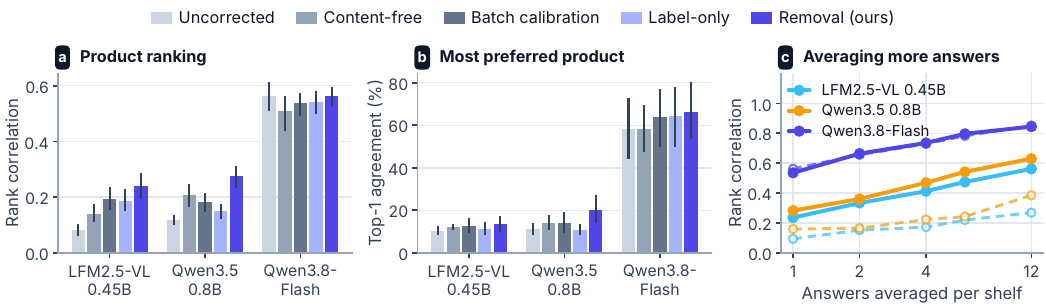}
\caption{Correction on held-out categories. (a) Rank correlation and (b) top-1 agreement of single-answer product utilities with the fitted product effects, for the uncorrected answer and four corrections. (c) Rank correlation against the number k of averaged answers, without (dashed) and with (solid) removal of the answer effects. Error bars are 95\% intervals.}
\label{fig:correction}
\end{figure*}

A scene is a shelf unit with four shelves and three positions per shelf. Each scene holds 12 products of one of 24 retail categories, each shown as the packshot of a distinct item from the TwinStore catalog \cite{twinstore2026} with its list price. Within a category, a cyclic Latin square of 12 arrangements moves the products so that every product stands at every position exactly once. Each product carries a letter badge from A to L \cite{yang2023setofmark}. The letters follow reading order, reverse order, or a random order per scene (\cref{fig:stimuli}). Because letters keep each answer to a single token, a model that exposes token probabilities gives the probability of every option in one query. Further renderings each change one aspect of the scene, namely schematic or blank packs, price tags at 0.8, 1, or 1.25 times the list price, a perspective camera, an upside-down, mirrored, or quarter-turned image with upright labels, the four shelves as separate images, the labels M to X, no labels with an answer by product name, and a text list. The protocol extends to 62 real shelf photos of SKU-110K \cite{goldman2019precise} and ShelfWise \cite{shelfwise2024products}, each with 6 to 12 labeled products and with masked and upside-down copies. Masking hides the products and keeps the layout and labels, exposing the answer effects of real images without product content. Reading tests ask for the letter of a named or priced product. A goal test replaces the category by a need that exactly one product meets. Each shelf yields a letter distribution from the open models and Qwen3.8-Flash and a single answer from Claude Haiku 5.5 and GPT-6 Luna.

\subsection{Factor Model and Correction}
\label{sec:method}
A decision $n$ offers options $j=1,\dots,J$. Option $j$ carries the letter $\ell_{nj}$, stands at the shelf position $c_{nj}$, and shows the product $k_{nj}$, and
\begin{equation}
P_n(j) = \frac{\exp\left(\alpha_{\ell_{nj}} + \gamma_{c_{nj}} + \delta_{k_{nj}}\right)}{\sum_{j'} \exp\left(\alpha_{\ell_{nj'}} + \gamma_{c_{nj'}} + \delta_{k_{nj'}}\right)},
\label{eq:factor}
\end{equation}
a conditional logit with a label effect $\alpha$ per letter, a layout effect $\gamma$ per position, and a product effect $\delta$ per product. The Latin square separates $\gamma$ from $\delta$, and the label orders separate $\alpha$ from $\gamma$. We fit \cref{eq:factor} by maximum a posteriori estimation with Gaussian priors, against the full letter distribution $p_n$ when available and against the single answer otherwise. Fitting the full distribution uses all 12 probabilities of a query, and every query informs all three effects. The difference of $\gamma$ between two groups of positions, such as top and bottom shelves, equals the change in the log-odds of a product against any other option when moved between the two positions. At the choice shares of a 12-product shelf, this change closely matches the log ratio of the product's choice shares at the two positions, the quantity that human shelf studies report \cite{chandon2009does,atalay2012shining}. On price-tag shelves a term $\beta \log \pi_{nj}$ enters the exponent, and $\beta(1-1/12)$ is the price elasticity at an even share. Rendering agreement is the reliability-corrected correlation of the two sets of product effects.

In \cref{eq:factor} the answer effects are shared by all scenes of an image type. Estimated once on calibration scenes of other categories, the answer effects give the product utility on a new shelf from one letter distribution, without training and without further queries:
\begin{equation}
\hat u_j = \log p_{nj} - \hat\alpha_{\ell_{nj}} - \hat\gamma_{c_{nj}}.
\label{eq:remove}
\end{equation}
We compare this removal with three corrections, namely label-only debiasing as in PriDe \cite{zheng2024large}, batch calibration \cite{zhou2024batch}, which needs all shelves of the test category, and content-free calibration with blank packs \cite{zhao2021calibrate}, which needs one extra query per scene. Removal takes out the fitted position effect, and an attention prior can add the position effect that people show. An attention map $A=(1-w)S+wG$ mixes spectral-residual saliency $S$ \cite{hou2007saliency} with a central Gaussian $G$ \cite{tatler2007central}, and with $a_j$ the mass of $A$ on product $j$,
\begin{equation}
q_{nj} \propto \exp(\hat u_j)\, a_j,
\label{eq:recompose}
\end{equation}
where the single weight $w$ is fitted on five human noticing effects \cite{chandon2009does} and the human choice effects of two studies \cite{chandon2009does,atalay2012shining} serve as a separate test. The prior shifts choices across shelf positions, while the product preferences still come from the model.

\begin{table}[t]
\centering
\footnotesize
\setlength{\tabcolsep}{2pt}
\begin{tabular}{@{}lccccc@{}}
\toprule
 & \multicolumn{3}{c}{Correct choice (\%)} & \multicolumn{2}{c}{Price elasticity} \\
\cmidrule(lr){2-4}\cmidrule(lr){5-6}
Model & Label & Price & Need & Shopper & Budget \\
\midrule
LFM2.5-VL 0.45B & 25 & 39 & 23 & $-0.02$ &  \\
 &  &  &  & {\scriptsize [$-0.06$, ${+}0.03$]} &  \\[1pt]
Qwen3.5 0.8B & 58 & 70 & 22 & $-0.03$ &  \\
 &  &  &  & {\scriptsize [$-0.10$, ${+}0.03$]} &  \\[1pt]
Qwen3.8-Flash & 94 & 100 & 96 & ${+}0.08$ & $-2.28$ \\
 &  &  &  & {\scriptsize [$-0.21$, ${+}0.33$]} & {\scriptsize [$-3.37$, $-1.78$]} \\[1pt]
Claude Haiku 5.5 & 94 & 99 & 95 & ${+}0.05$ & $-2.69$ \\
 &  &  &  & {\scriptsize [$-0.40$, ${+}0.64$]} & {\scriptsize [$-4.20$, $-2.18$]} \\[1pt]
GPT-6 Luna & 93 & 100 & 90 & ${+}0.36$ & $-5.64$ \\
 &  &  &  & {\scriptsize [$-0.11$, ${+}0.80$]} & {\scriptsize [$-10.44$, $-4.69$]} \\[1pt]
\bottomrule
\end{tabular}
\caption{Shelf reading, goal-directed choice, and price response. \emph{Label}, \emph{Price}, and \emph{Need} give correct choices in percent (chance 8\%), and the last two columns give price elasticities for the shopper and the budget-shopper prompts with 95\% intervals below.}
\label{tab:shopper}
\end{table}

\subsection{Results}
\label{sec:shelf-results}
We study two open models, LFM2.5-VL 0.45B \cite{amini2025lfm2} and Qwen3.5 0.8B \cite{qwen2026qwen35}, and the three hosted models of \cref{sec:shopper}, which answer without a reasoning step.

\paragraph{Reading and goals (R1, R2).}
Because each reading question asks for one letter, a correct answer requires finding the product or the price in the image and then the adjacent label. The hosted models answer correctly for a named product on more than 90\% of front views and for a priced product on nearly every shelf, and read the bottom shelf about as well as the top one (\cref{tab:shopper}). The top-of-image preference reported below arises in choosing and not in reading. Given a specific need, the hosted models choose the one product that meets the need on 90\% to 96\% of shelves, far above chance, and keep choosing that product at every position and label. A clear goal thus outweighs both answer effects.

\paragraph{Product preferences (R3, R4).}
The hosted models keep their product preferences across scene changes, with corrected correlations near one for the perspective view, the labels M to X, and paraphrased prompts. Under the mirrored view and the quarter turn, which reverse or rotate the printed name on every pack, the preferences also hold and therefore rest on the products, not on the orientation of the names. Product effects on photo shelves agree with those of a text list at only 0.56 to 0.60, evidence that pack images carry preferences beyond product names, information a text-only respondent lacks. As budget shoppers, the hosted models buy less of a product when its price rises (\cref{tab:shopper}), Qwen3.8-Flash and Claude Haiku 5.5 close to the mean elasticity of $-2.62$ for brand sales \cite{bijmolt2005new} and GPT-6 Luna still more strongly. Without a budget the models choose by product despite reading every price. The stated profile thus sets the price response, as in \cref{sec:real}, where all shoppers have budgets.

\begin{table}[t]
\centering
\small
\setlength{\tabcolsep}{2pt}
\resizebox{\ifdim\width>\linewidth\linewidth\else\width\fi}{!}{%
\begin{tabular}{lccc}
\toprule
Model & Letter A & Top vs bottom shelf & Center vs sides \\
\midrule
LFM2.5-VL 0.45B & ${+}0.92$ {\scriptsize [${+}0.88$, ${+}0.95$]} & $-0.32$ {\scriptsize [$-0.42$, $-0.23$]} & ${+}0.06$ {\scriptsize [${+}0.01$, ${+}0.10$]} \\
Qwen3.5 0.8B & ${+}0.84$ {\scriptsize [${+}0.79$, ${+}0.91$]} & $-0.54$ {\scriptsize [$-0.62$, $-0.46$]} & $-0.56$ {\scriptsize [$-0.63$, $-0.50$]} \\
Qwen3.8-Flash & ${+}0.84$ {\scriptsize [${+}0.63$, ${+}1.05$]} & ${+}0.66$ {\scriptsize [${+}0.39$, ${+}0.91$]} & ${+}0.09$ {\scriptsize [$-0.15$, ${+}0.32$]} \\
Claude Haiku 5.5 & ${+}1.24$ {\scriptsize [${+}0.89$, ${+}1.89$]} & ${+}1.18$ {\scriptsize [${+}0.91$, ${+}1.77$]} & ${+}0.03$ {\scriptsize [$-0.19$, ${+}0.29$]} \\
GPT-6 Luna & ${+}2.17$ {\scriptsize [${+}1.57$, ${+}3.08$]} & ${+}0.68$ {\scriptsize [${+}0.02$, ${+}1.55$]} & ${+}0.30$ {\scriptsize [$-0.06$, ${+}0.63$]} \\
\midrule
Human shoppers &  & ${+}0.11$ & ${+}0.16$ to ${+}0.50$ \\
\bottomrule
\end{tabular}}
\caption{Label and layout effects on product-photo shelves, pooled over label orders, in log-odds with 95\% intervals. The human row gives the choice effects, as log ratios of choice shares, of shoppers choosing from shelf photographs and on-screen displays \cite{chandon2009does,atalay2012shining}.}
\label{tab:effects}
\end{table}

\paragraph{Answer effects.}
The hosted models favor the letter A, which multiplies the odds of a choice by 2.3 to 8.8 relative to an average letter at the same position (\cref{tab:effects}). The effect belongs to the letter itself and not to the first label of a set, since the first letter M of the labels M to X gains far less. The hosted models also favor the top shelf over the bottom shelf by 0.66 to 1.18 log-odds, compared with 0.11 for people choosing from shelf photographs, the same task that \bench poses. This layout effect follows the image and not the shelf. When the image is turned upside down, Qwen3.8-Flash and Claude Haiku 5.5 favor the physical bottom shelf, which then lies at the top of the image, and one factor model over four views assigns almost all of the vertical advantage of these two models to the top of the image and almost none to the physical top shelf. When the four shelves come as separate images in a random order, the advantage of Claude Haiku 5.5 and Qwen3.8-Flash moves to the shelf shown first, which ties the advantage to the order of the input. The shelf-height preference of \cref{sec:real} is thus a preference for the top of the image, several times stronger than that of people. Both effects come from where an option appears and which letter the option carries, and not from the products. A study can therefore remove both effects and add back the position effect of people through the attention prior.

\paragraph{Removing the answer effects (R5).}
The answer effects are shared across scenes, and interactions with the label order or the shelf barely improve the prediction of held-out answers. Effects estimated on other categories therefore recover the model's product preferences on a new category, given by the product effects fitted on the other renderings of that category (\cref{fig:correction}). For Claude Haiku 5.5 and GPT-6 Luna, which give one answer per shelf, the correction and counting use the same 12 answers per shelf and keep the products that no answer chose tied. The rank correlation then rises from 0.44 to 0.54 and 0.52, respectively, because the letter A and the top of the image no longer count as preference. For Qwen3.8-Flash, whose single answer already ranks the products well, removal keeps this ranking and more often agrees with the reference on the most preferred product, whereas the other corrections lower the ranking. Label-only debiasing leaves the layout effect in place, and content-free calibration estimates the answer effects on blank packs, where the answer effects differ from those on product photos. For the two open models, removal more than doubles the rank correlation of a single answer and outperforms the three other corrections against this reference and against a reference built without the factor model, with every paired interval excluding zero. Masked copies share the layout and the labels of the photos, and effects estimated on masked copies of other photos lower the share of the first letter for Qwen3.8-Flash on real photos from 37.3\% to 10.0\%, near the 9.0\% of an even choice, and on the masked copies from 45.3\% to 13.9\%. A study of real shelves can calibrate on masked copies of the same photos. The attention prior adds a center effect approaching that of people for models with letter probabilities, while Qwen3.8-Flash keeps the ranking.

\section{Conclusion}
\label{sec:conclusion}

VLM shoppers that perceive a store through images follow real consumers. Built from household records, the shoppers hit held-out purchases as often as the households' own past trips, infer group and market differences without fitting, and respond to situations, prices, and displays within measured ranges. \bench and its factor model separate VLM product preferences from two answer effects, a preference for the letter A and for positions in the image, which explain the shelf-height preference in the store and are removed with one estimate per image type. The factor model applies to any VLM that selects among marked image regions, as web agents do. Synthetic shopper studies can thus take real purchase statistics as persona inputs, vary labels and positions, and remove the answer effects before reading product preferences, letting a retailer build synthetic shoppers from its own records and shelf images.

\section*{Acknowledgments}
This work was supported by the Startup Growth Technology Development Program (Deep-Tech TIPS) of the Ministry of SMEs and Startups (MSS, Korea) (Project No.~RS-2025-25461644).

\section*{LLM Usage Statement}

We used Claude (Anthropic) through Claude Code to draft and edit the paper and to write the code for the experiments, analyses, figures, and tables, and ChatGPT (OpenAI) to review drafts. The VLMs named in \cref{sec:shopper,sec:shelf-results} are the subjects of the experiments.

{
    \small
    \bibliographystyle{ieeenat_fullname}
    \bibliography{refs}
}

\clearpage
\appendix
\begin{center}
{\Large\bfseries Appendix}
\end{center}
\vspace{0.5em}
The appendix first gives the experiment code (\cref{sec:suppl-code}). It then details the VLM synthetic shoppers, the data and protocol, and further results on real purchases (\cref{sec:suppl-shopper,sec:suppl-protocol,sec:suppl-real-results}), describes \bench, its renderings and prompts, and the factor model (\cref{sec:suppl-bench,sec:suppl-design,sec:suppl-model}), and gives the evidence for each requirement, the analysis of the answer effects, and their removal (\cref{sec:suppl-evidence,sec:suppl-answer-effects,sec:suppl-correction}).

\section{Experiment Code}
\label{sec:suppl-code}

The experiment code is available at
\begin{center}
\url{https://anonymous.4open.science/r/seeing-like-a-shopper-0D27/}
\end{center}
The code renders the \bench shelves from the product catalog and packshots of TwinStore \cite{twinstore2026}, queries the open and hosted models, fits the factor model, removes the answer effects, and builds the attention priors, and also runs the purchase-level simulations on the public retail datasets together with the analyses of these simulations. The repository contains the experiment code only and no data or results. The datasets are obtained from the original sources, and the repository describes where each dataset is placed and which script produces each result.

\section{VLM Synthetic Shoppers in Detail}
\label{sec:suppl-shopper}

\begin{table}[!htbp]
\centering
\small
\resizebox{\linewidth}{!}{%
\begin{tabular}{@{}llll@{}}
\toprule
 & Own records & Age group & Text-only persona \\
\midrule
Demographics and home location & given & given & given \\
Visit day and time, items and categories per trip & the household's & a household of the same age group & a random household \\
Purchase rate of each category & the household's & age-group average & store average \\
Usual product of each category & given & not given & not given \\
\bottomrule
\end{tabular}}
\caption{Information in the persona under each condition.}
\label{tab:suppl-persona}
\end{table}

\paragraph{Personas.}
In The Complete Journey, each persona corresponds to one household and is written from records before the test period, as \cref{tab:suppl-persona} lists for each condition. In every condition the model decides whether to enter, which categories, products, and quantities to buy, which product on a shelf the shopper notices first, and how interested the shopper is. The numbers of items and categories per trip come from one past trip of the household itself, of a household of the same age group, or of a random household, and are given as inputs. The evaluation in The Complete Journey thus focuses on what the shopper buys and how the shopper responds to changes. Before each trip, the categories that the household needs are sampled from the household's purchase rates and listed as, for example, ``needed this time: yogurt (usually brand A or brand B)''. The model may follow the list, remove items from the list, or add categories, and added categories are bought as well. The suggested categories anchor the trip in the purchase rates of the persona, and the added and removed categories show what the model infers about the household. In the text-only condition, the persona keeps only the store-average rates and the sentences that describe the household.

\paragraph{The store.}
The store holds the 30 categories with the highest sales in the first 26 weeks of The Complete Journey and eight products in each, 240 products in all, at representative prices from the same weeks. Each category stands on one shelf, and within a category the best-selling products occupy the eye-level, waist-level, top, and bottom shelves in that order. The directory is a map of the shelves labeled with category names, on which promoted shelves carry a red ribbon. The model plans the route from the directory as a shopper does at the entrance. The product list that accompanies the directory appears in a new random order each time, which keeps the model from favoring the first products listed. Shelf images print the product name on each pack and show a yellow tag with a number and the price, and promoted products carry a ribbon. Because the same store is used for every country, differences between groups come from the shoppers.

\paragraph{Decisions.}
At the entrance the model receives the persona, the time of the visit, the situation of the day, a budget, the directory image, and the product list, and answers in one step with the purpose of the visit, whether to enter, and the categories, product names, and quantities that the shopper plans to buy. Each trip keeps the trip size given as input, and the household's rates supply any categories beyond the plan. While walking along the route, the shopper views up to three nearby shelves. For each shelf image the model names the product that the shopper notices first and rates the shopper's interest in that product from 0 to 10, and the product is bought with probability $s \times \text{interest}/10 \times h$, with $s=0.5$ and $h$ equal to 1.00 at eye level, 0.85 at waist level, 0.70 on the top shelf, and 0.55 on the bottom shelf. The model decides which product attracts interest and how much, and this rule converts interest into purchases at the same rate for every shopper. When a plan names a category without a product, the promoted products of the category receive a weight of 1.8 in the choice, the fixed-weight path of a promotion in \cref{fig:suppl-markets}(c). A household receives the same budget and maximum number of items whether a VLM or the rule-based shopper makes the trip.

\paragraph{Situations and models.}
Weekday, season, events such as a major sports broadcast or the year-end holidays, and the time of the visit enter the simulation in two ways. The model reads descriptions such as ``summer, 30\,$^\circ$C'' or ``Sunday with a big game'', and the sampling of needed categories can be weighted by situational responses measured in real purchase records. The situational effects for descriptions alone use only the first path. Each shopper is assigned at random to one of Claude Haiku 5.5, Qwen3.8-Flash, and GPT-6 Luna, and the results describe this mixed population unless a model is named (\cref{tab:suppl-models}).

\section{Data and Protocol}
\label{sec:suppl-protocol}

\paragraph{Held-out households.}
The year of The Complete Journey is split into three periods. The store and the personas are built from weeks 1 to 26, and the simulator is calibrated on 150 households in weeks 27 to 39 that take no part in the evaluation. Two panels of 150 households each are sampled independently from the remaining 651 households, 268 different households in all. The personas of these households are written from weeks 1 to 39, and the synthetic purchases are compared with the real purchases of weeks 40 to 53. Because households differ in the number of trips, each real trip is weighted by the inverse of the number of trips of the same household. Every household thus carries the same weight, and frequent shoppers do not dominate the comparison. For differences between groups, each household makes five trips at different times and with different trip sizes, which reduces the noise of household-level estimates.

\paragraph{Unseen markets.}
For X5 RetailHero and Ta-Feng, 30 customers are sampled from each of 12 groups, defined by age and gender or by age and residence, which gives 360 synthetic shoppers per dataset. The store and the simulator of The Complete Journey carry over unchanged. Because the persona holds only the description of the customer, the model also decides the size of each trip. An effect counts as reproduced when the correlation with the real effect is positive at permutation $p<0.05$, and the ratio of items per trip between far and near customers counts as reproduced when the interval of the ratio lies above one and the ratio lies closer to the real ratio than the transferred value.

\paragraph{Baselines.}
\emph{Past trips} uses a real trip of the same household before the test period as the prediction. The \emph{rate list} builds a basket of the same size from the same purchase rates without a language model, with the household's own rates and usual products in the own-records condition and the age-group rates in the age-group condition. \emph{Learned} models predict category rates from seven demographic attributes by ridge regression or $k$-nearest neighbors (kNN) with $k=25$, trained on 383 households outside the held-out panels and the calibration households. The \emph{rule-based} shopper receives the same persona, selects products from a category list fixed when the persona is written, decides unplanned purchases at the shelf with fixed probabilities, and responds to price through a fixed elasticity term. A model \emph{transferred} from the US data applies the age gradient estimated on the US data to the other markets. Because the rule-based shopper shares the simulator, the persona, and the store with the VLM shopper, the comparison between the two isolates the decisions that the language model makes.

\paragraph{Measures.}
\emph{Basket mix} is the Jensen--Shannon divergence (JSD) between the shares of the 30 categories in synthetic and real purchases, which is zero for identical distributions. The measure compares the mix of what is bought, independent of how many items a trip holds. For the \emph{hit rate} of a household, we take for each category in the synthetic basket of the household the share of real test-period trips that contain the category, and average the shares. Baskets sampled by store-average rates give the reference without personalization. The \emph{classifier test} \cite{lopezpaz2017revisiting} trains gradient boosting or logistic regression to distinguish a real trip of a household from a trip of the same household made by a shopper or baseline. The classifier features are whether each of the 30 categories is bought, spending, time, weekday, age group, and items and units per trip, optionally with the share of products the household bought before, the store-brand share, and the sales rank of the products. We report the 5-fold cross-validated area under the ROC curve (AUC) next to the AUC between two real trips. An AUC of 0.5 means that the classifier cannot tell a synthetic trip from a real one, and the AUC between two real trips sets the level that real variation alone reaches. For \emph{differences between groups}, we compute how far each demographic group's share of each category deviates from the mean, in real and synthetic purchases, and correlate the two sets of deviations, with significance from 1,000 random reassignments of the synthetic groups. Correlating deviations instead of shares removes what all groups have in common, and the measure thereby rewards only the differences between groups. \emph{Situational effects} are the ratios of purchase rate and spending under six situations (weekends, days with major sports broadcasts, summer, winter, the year-end season, and evening visits) to a reference situation for eight categories and total spending, 54 effects in all, compared by the correlation of the logarithms.

\section{Additional Results on Real Purchases}
\label{sec:suppl-real-results}

\begin{table}[!hb]
\centering
\small
\resizebox{\linewidth}{!}{%
\begin{tabular}{@{}llccl@{}}
\toprule
Measure & Real or reference & VLM & Rule-based & Baseline without a VLM \\
\midrule
Basket mix, JSD, own records & past trips 0.009 & 0.011 & 0.037 & rate list 0.009 \\
Hit rate, own records & past trips 0.288 & 0.291 & 0.263 & rate list 0.307 \\
Group differences $r$, age-group rates & past trips 0.52 & 0.28 & 0.11 & age-group list 0.16 \\
Real purchase rate of categories added beyond the list & same base rate 0.136 & 0.166 & & \\
Store-brand share, low / high income & 1.16 & 1.10 to 1.12 & 0.96 to 1.00 & \\
Instacart next-order aisle hit & order history 0.56 & 0.52 & 0.38 & popular aisles 0.41 \\
Repurchase share & 0.70 & 0.62 & 0.69 & rate list 0.53 \\
X5 alcohol trips, $r$ over 12 groups & criterion $p<0.05$ & 0.56 & 0.30 & US transfer 0.03 \\
X5 items per trip, $r$ & criterion $p<0.05$ & 0.20 & $-0.07$ & US transfer 0.77 \\
X5 own-brand share, $r$ & criterion $p<0.05$ & $-0.81$ & $-0.67$ & US transfer 0.28 \\
Ta-Feng items per trip, far / near & 1.33 & 1.30 & 1.06 & US transfer 1.00 \\
54 situational effects $r$, descriptions only & & 0.43 & no response & \\
54 situational effects $r$, with measured responses & & 0.72 & 0.90 & \\
Own-price elasticity, products & $-0.70$ to $-2.62$ & $-1.09$ & $-0.87$ & \\
Own-price elasticity, categories & $-0.27$ to $-0.81$ & $-0.21$ & $-0.57$ & \\
Display multiplier & 1.37 to 1.54 & 1.40 & & \\
\bottomrule
\end{tabular}}
\caption{Summary of the comparisons with real purchases. VLM denotes the mixed population of the three hosted models, and the last column gives the strongest baseline without a VLM that uses the same information.}
\label{tab:suppl-summary}
\end{table}

\Cref{tab:suppl-summary} collects the comparisons with real data. With the household's own records the VLM shopper reaches the hit rate of the household's own past trips and comes close to past trips in basket mix. Beyond what purchase rates provide, the VLM shopper reproduces differences between groups described only in words, adds categories that the household buys more often than categories of the same base rate, carries effects of gender and travel distance into markets without fitting to these markets, responds to situations given only as descriptions, and reads the shelf image. Group differences are reproduced best when age-group rates are given, and the agreement with the 54 situational effects rises to $r=0.72$ when measured responses are given.

\begin{figure}[!htbp]
\centering
\includegraphics[width=\linewidth]{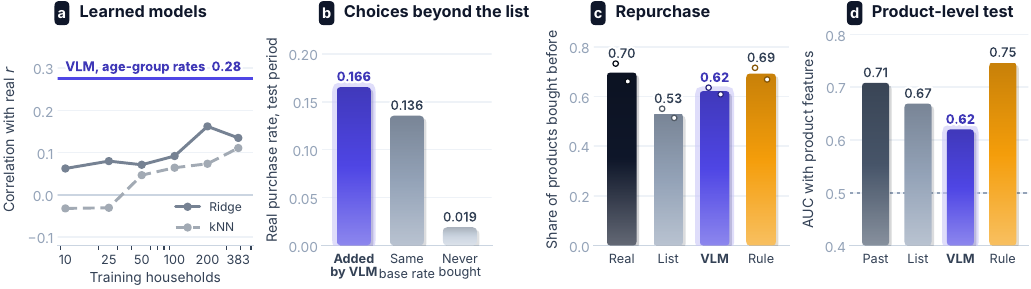}
\caption{Further household results. (a) Correlation of group differences for ridge regression and k-nearest-neighbor (kNN) models trained on increasing numbers of households, against VLM shoppers given age-group rates. (b) Real test-period purchase rate of categories that VLM shoppers add beyond their need list, of categories with the same base rate, and of categories the household never bought. (c) Share of basket products that the household bought before the test period. (d) Logistic-regression AUC with product-level features added (previously bought products, store-brand share, sales rank); for past trips the trip itself counts as previously bought.}
\label{fig:suppl-households}
\end{figure}

\paragraph{Households and groups.}
Ridge regression and $k$-nearest neighbors do not reach the correlation of the VLM shoppers with age-group rates even when trained on all 383 households (\cref{fig:suppl-households}(a)). Categories that VLM shoppers add beyond the need list in the age-group condition are bought in the real test period at 0.166, against 0.136 for categories with the same base rate and 0.019 for categories the household never bought (\cref{fig:suppl-households}(b)), with differences of $+0.025$ and $+0.027$ in the two panels, both with intervals excluding zero. VLM shoppers select their usual products depending on the context, which brings the repurchase share of VLM shoppers closer to the real one than that of the rate list built from the same records (\cref{fig:suppl-households}(c)). With product-level features, the classifier distinguishes VLM trips from real ones with a lower AUC than those of any baseline (\cref{fig:suppl-households}(d)). In Instacart online orders of 141 users, VLM shoppers hit the aisles of the next order at 0.52, near the users' own order history and above popular aisles at 0.41, while the rule-based shopper reaches 0.38.

\begin{figure}[!htbp]
\centering
\includegraphics[width=\linewidth]{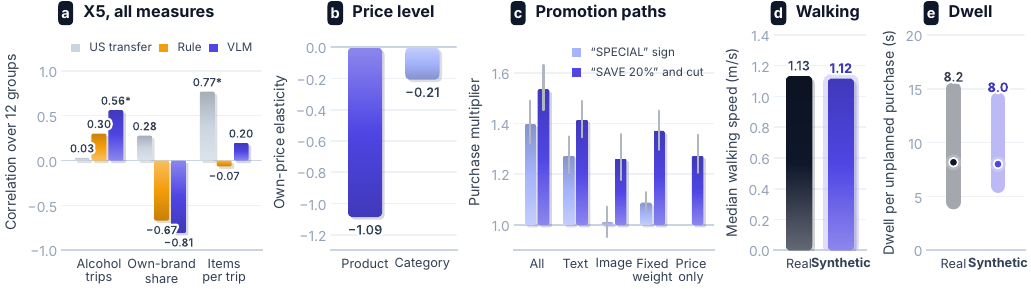}
\caption{Further market, price, promotion, and movement results. (a) Correlation with X5 across 12 age and gender groups for the share of trips with alcohol, the own-brand share, and items per trip; an asterisk marks permutation $p<0.05$. (b) Own-price elasticity when half of the products or half of the categories change price by 20\%. (c) Purchase multiplier of promoted products when the promotion reaches the shopper through every path, only the text, only the shelf image, only a fixed weight, or only the price, with 95\% intervals. (d) Median walking speed against 13,159 tracks in the ATC shopping center. (e) Median dwell time per unplanned purchase with its interquartile range against mobile eye tracking in a supermarket.}
\label{fig:suppl-markets}
\end{figure}

\paragraph{Markets, price, promotion, and movement.}
In X5, the VLM shoppers reproduce the alcohol pattern by age and gender, which the transferred US model, lacking a gender term, misses (\cref{fig:suppl-markets}(a)). A price change of whole categories changes the purchases of VLM shoppers far less than a price change of products (\cref{fig:suppl-markets}(b)), as in measured food demand \cite{andreyeva2010impact}. The rule-based shopper shares the budget, the store, and the trip sizes with the VLM shopper and loses nearly all of the price response once the fixed elasticity term is removed, whereas VLM shoppers keep most of the response without the fixed elasticity term of the simulator. The price response of VLM shoppers therefore comes from the model and not from the budget. Isolating each path by which a promotion reaches the shopper shows that the model's reading of the promotion text drives the response (\cref{fig:suppl-markets}(c)). Walking speed and dwell time are calibrated to the ATC tracks and the eye-tracking study, and the synthetic shoppers walk at a median 1.12\,m/s against 1.13\,m/s and dwell a median 8.0\,s per unplanned purchase against 8.2\,s (\cref{fig:suppl-markets}(d, e)).

\begin{table}[!htbp]
\centering
\small
\resizebox{\linewidth}{!}{%
\begin{tabular}{@{}lcccc@{}}
\toprule
 & Instacart aisle hit & Bottom shelf, image & Bottom shelf, text & Snack trips (\%): drinks / far end / beer \\
\midrule
Claude Haiku 5.5 & 0.54 & 0.24 & 0.24 & 0.3 / 0.0 / 0.3 \\
Qwen3.8-Flash & 0.50 & 0.14 & 0.21 & 0.7 / 0.0 / 1.0 \\
GPT-6 Luna & 0.53 & 0.15 & 0.21 & 0.3 / 0.0 / 2.0 \\
Rule-based & & & & 3.7 / 0.3 / 2.7 \\
\bottomrule
\end{tabular}}
\caption{Results of each hosted model on its own. Instacart aisle hit of the next order; share of first-noticed products on the bottom shelf when products rotate through the four levels, against 0.25 for no preference; and the share of trips with an unplanned snack purchase when the snack shelf stands next to soft drinks, at the far end of the aisle, or in front of the beer cooler (300 trips each).}
\label{tab:suppl-models}
\end{table}

\paragraph{Models and layout.}
\Cref{tab:suppl-models} gives the results of each model on its own. Qwen3.8-Flash and GPT-6 Luna notice the bottom shelf less with the shelf image than with the same shelf as text. Moving the snack shelf to the far end of the aisle reduces unplanned snack purchases to zero in 300 trips for each of the three models, and placing the snack shelf in front of the beer cooler raises these purchases for Qwen3.8-Flash and GPT-6 Luna, in the direction of measured effects of placing complementary categories together \cite{bezawada2009cross}.

\section{\bench in Detail}
\label{sec:suppl-bench}

\begin{figure}[!htbp]
\centering
\includegraphics[width=\linewidth]{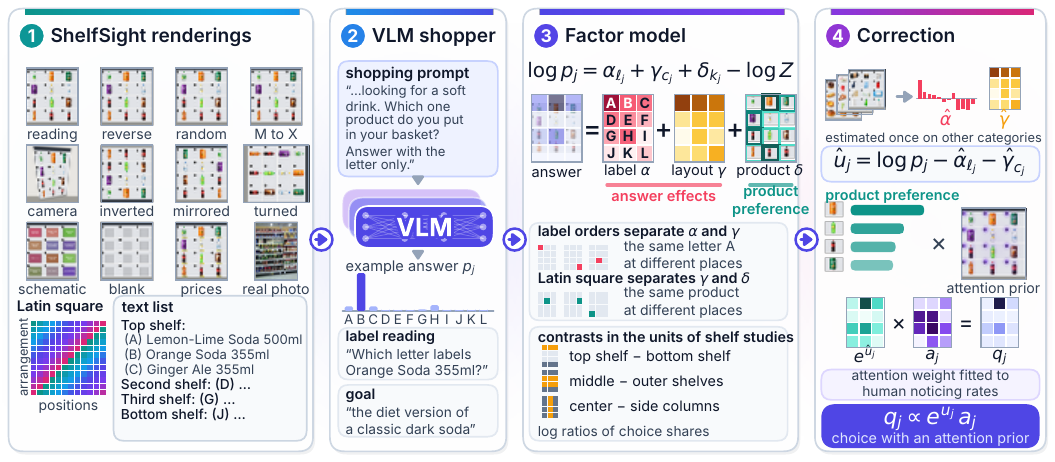}
\caption{Overview of ShelfSight and the factor model on one example shelf. (1) Renderings of the shelf. (2) A VLM shopper with its prompt, an example answer, and the reading and goal tests. (3) The factor model, which divides the log probability of each option into label, layout, and product effects; Z normalizes over the options. (4) The correction and the attention prior.}
\label{fig:suppl-overview}
\end{figure}

\Cref{fig:suppl-overview} illustrates \bench and the factor model on one shelf. \bench poses to a model the choice that a shopper makes at a shelf, which product to put in the basket. Products, positions, and labels vary independently across renderings, and each rendering changes one aspect of a base scene. Every decision is a choice among 12 labeled products, and the model answers with a single letter, or with a product name in the label-free view.

\paragraph{Scenes and placement.}
A scene is a shelf unit with four shelves and three positions per shelf. Each scene holds 12 products of one of 24 retail categories, such as soft drinks, bakery goods, or garden supplies, each shown as the packshot of a distinct item from the product catalog of TwinStore \cite{twinstore2026}, which also gives the list price of each item. Within a category, a cyclic Latin square of 12 arrangements places product $(k+s) \bmod 12$ at position $k$ in arrangement $s$, and every product thereby stands at every position exactly once.
\Cref{tab:suppl-bench-stats} summarizes the size of the benchmark.

\begin{table}[!htbp]
\centering
\small
\begin{tabular}{@{}ll@{}}
\toprule
Element & Size \\
\midrule
Retail categories & 24, with 12 products each \\
Shelf unit & 4 shelves $\times$ 3 positions \\
Arrangements per category & 12, from a cyclic Latin square \\
Label orders & reading, reverse, and random; letters A to L or M to X \\
Shelves per rendering & 144 with all arrangements, 72 with every second (12 categories) \\
Salience scenes & 48 shelves per manipulation \\
Reading questions & 72 per view for names, 70 for prices \\
Goal-directed needs & 24, two in each of 12 categories \\
Real shelf photos & 62 (48 SKU-110K, 14 ShelfWise), 124 shelves per version \\
Models & 2 open and 3 hosted VLMs \\
\bottomrule
\end{tabular}
\caption{\bench at a glance.}
\label{tab:suppl-bench-stats}
\end{table}

\paragraph{Labels.}
Each product carries a letter badge from A to L at the top-left corner of the product's place, in the style of set-of-mark prompting \cite{yang2023setofmark}. Because letters keep each answer to a single token, one query gives the probability of every option for a model that exposes token probabilities. In reading order A labels the top-left product and the letters run left to right and top to bottom, in reverse order A labels the bottom-right product, and in random order the letters are permuted at random once per scene. Reading and reverse order link each letter to a position in opposite directions and place the same letters at opposite corners. A letter effect and a corner effect therefore cannot both explain the choices in both orders. Random order changes this link from scene to scene.

\paragraph{Renderings.}
The front view shows the unit from the front with product photos (Fig.~\ref{fig:stimuli}(a)). Four variants change what the products look like or cost. Schematic packs print the product name on a colored box, and blank packs show the same plain gray pack everywhere (Fig.~\ref{fig:stimuli}(d)), and any remaining preference on blank packs comes from labels and positions. Price tags show the list price times 0.8, 1, or 1.25, chosen at random for every product and placement, which makes price unrelated to position and label. A dominance variant shows one product twice side by side, at 0.8 and 1.25 times the list price of that product. Four variants change the viewpoint. The perspective view warps the front view to a camera at a shopper's eye height (Fig.~\ref{fig:stimuli}(b)), the upside-down view inverts the front view vertically (Fig.~\ref{fig:stimuli}(c)), the mirrored view reverses the image left to right, and the quarter turn rotates the image clockwise and turns the top shelf into the right column, all with upright labels. A further variant shows the four shelves as separate images in a random order. Three variants change the answer. A second label set uses the letters M to X, which avoids the letter A, the label-free view asks for the product name, and the text rendering lists the shelf one line per level, as in ``Top shelf: (A) name, (B) name, (C) name'', from the top shelf down or from the bottom shelf up. Each variant keeps every other aspect of the base scene, and a change in the fitted effects can thereby be traced to the one aspect that the variant changes.

\paragraph{Real shelf photos.}
We apply the protocol to 48 test images of SKU-110K \cite{goldman2019precise} and 14 images of the ShelfWise products dataset \cite{shelfwise2024products}, both annotated with product bounding boxes. One product in each region of a four-by-three grid receives a label (Fig.~\ref{fig:stimuli}(e)), and a masked copy paints the labeled boxes gray (Fig.~\ref{fig:stimuli}(f)). In a photo the position and the content of a product are tied, and upside-down copies with upright labels separate position from content.

\paragraph{Salience, reading, and goals.}
Salience scenes show one target product in the middle of the second or third shelf, either plain or with a red corner ribbon with or without a ``SALE'' bar, a yellow frame, a pack approximately 30\% larger, or a second facing in place of the neighboring product, while the remainder of the scene is unchanged. For the second facing, the plain scene leaves out the product that the second facing displaces. Two reading tests ask which letter labels a named product, or the product with a given price, on the top or bottom shelf. A goal-directed test replaces the category in the prompt by a need that exactly one product meets, described without the product name, such as ``something to check whether you have a fever''.

\paragraph{Prompts and answers.}
The shopping prompt states the category the shopper is looking for, such as ``a soft drink'', says that every product has a letter label, and asks which one product goes into the basket, answered with the letter only. Two paraphrases test whether the results depend on the wording of the prompt. The open models give the probability of each letter. Qwen3.8-Flash gives the probabilities of its five most likely answers, and the remaining probability is shared evenly by the other letters. Claude Haiku 5.5 and GPT-6 Luna give one answer per shelf.

\paragraph{Human references.}
The human references come from two studies in which people chose from shelf images \cite{chandon2009does,atalay2012shining}. As log ratios of choice shares, the two studies give a center effect of $+0.16$ to $+0.50$ and a top-bottom effect of $+0.11$ (\cref{sec:suppl-human}).

\section{Renderings and Prompts}
\label{sec:suppl-design}

\subsection{Renderings}
\label{sec:suppl-renderings}

\begin{figure}[!htbp]
\centering
\includegraphics[width=\linewidth]{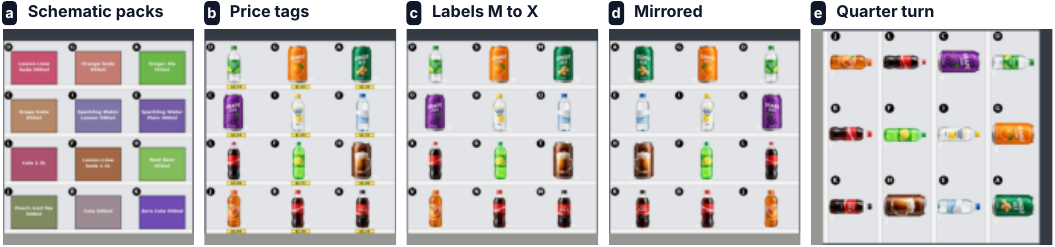}
\caption{Further renderings of the soft-drink shelf. (a) Schematic packs that print each product name on a colored box. (b) Product photos with price tags at 0.8, 1, or 1.25 times the list price. (c) The label set M to X. (d) The front view mirrored left to right. (e) The front view turned a quarter clockwise, with the top shelf on the right. Labels stay upright in every view.}
\label{fig:stimuli-more}
\end{figure}

\paragraph{Categories and products.}
The 24 categories and the need stated in the shopping prompt are soft drinks (a soft drink), beauty products (a beauty product), bakery (something from the bakery), ice cream (an ice cream), household items (a household item), fresh produce (fresh fruit or vegetables), personal care (a personal care item), hardware (a hardware item), gardening (a gardening item), plant-based food (a plant-based food), breakfast spreads (a breakfast spread), coffee and tea (coffee or tea for home), bath and body (a bath or body product), first aid (a first aid item), supplements (a nutrition supplement), pet supplies (something for your pet), snacks (a snack), drinks (a drink), baking ingredients (a baking ingredient), frozen meals (a frozen meal), baby care (a baby care product), oral care (an oral care product), ready meals (a ready meal), and music accessories (a music accessory). Each category holds 12 distinct generic products, each with a packshot and a list price from the product catalog of TwinStore \cite{twinstore2026}. The soft drinks, for example, include cola, zero cola, lemon-lime soda, ginger ale, sparkling water, and root beer in several sizes. The first 12 categories form the core set of every rendering. For LFM2.5-VL 0.45B, the product-photo renderings in all three label orders, the perspective view, the blank packs, and the text lists extend to all 24 categories.

\paragraph{Placements.}
A rendering shows each category in all 12 arrangements of the Latin square, which gives 144 shelves per model with 12 categories, or in every second arrangement, which gives 72. With every second arrangement a product appears at six of the 12 positions, at least one on each of the four shelves, and every position still holds six different products. The mirrored and quarter-turn views, the paraphrased prompts, and the goal-directed test use every second arrangement. The salience scenes use two arrangements per category, each shown once with the target in the middle of the second shelf and once in the middle of the third, which gives 48 shelves per manipulation. The label-reading test asks 72 questions per view and the price-reading test 70.

\paragraph{Labels.}
Because the random order is fixed for a scene, the upside-down, mirrored, turned, price-tag, and separate-image renderings carry the same letters on the same products as the front view with random labels. The labels M to X keep the same assignment of labels to positions, and differences between the two label sets therefore come from the letters themselves.

\paragraph{Price tags.}
A white tag under each product shows the price of the product. Each product's price is 0.8, 1, or 1.25 times the list price, chosen at random for each scene independently of position and label and kept the same in every price rendering of that scene. The shopper without a budget and the shopper on a tight budget see identical shelves and differ only in the prompt. In the dominance scenes a second copy of the target, which stands in the middle of the second or third shelf, replaces the neighboring product on the same shelf and receives that product's letter. The two copies carry 0.8 and 1.25 times the list price, and the cheaper copy alternates between arrangements to keep both sides of the pair balanced.

\paragraph{Viewpoints.}
The perspective view warps the front image to a camera at the eye height of a shopper standing in front of the unit and slightly to the left, which makes the shelves converge and the packs on the right appear narrower. The upside-down view turns the image over vertically, the mirrored view reverses the image left to right, and the quarter turn rotates the image clockwise by 90 degrees. All three redraw the badges upright at the top-left corner of each product's place in the transformed image, and every letter reads as on the front view. Each view keeps the products and the labels of the front view and changes only the geometry, which tests whether preferences follow the products or the pixels.

\paragraph{Shelves as separate images.}
The separate-image rendering cuts the front view with random labels into the four shelves and presents the shelves to the model as four images in a random order, with a prompt that says the images come in no particular order. Two crops are used. Crops with shelf edges are cut at the edges of each shelf, and the dark top edge of the unit then remains on the image of the top shelf and marks that shelf. Crops without shelf edges are cut inside the edges, and no edge marks the top shelf. Qwen3.8-Flash sees both crops, which isolates the role of the edge within one model, and Claude Haiku 5.5 and GPT-6 Luna see the crops with and without shelf edges, respectively. Because separate images break the spatial layout of the unit, a remaining preference for one shelf can come only from the order of the images or from cues on the shelves themselves.

\paragraph{Text lists.}
The text rendering lists the shelf as four lines, one per level, in the form ``Top shelf: (A) name, (B) name, (C) name'', followed by the second shelf, the third shelf, and the bottom shelf. The bottom-first version lists the same lines in the opposite order. The text list keeps the products and the labels and removes the image, which isolates the contribution of the image to the preferences.

\paragraph{Real shelf photos.}
The four rows of the grid are quarters of the image height and thus measure position in the image. In each region the largest annotated box whose center falls inside the region receives the label, and a region without an annotated box receives none. A photo therefore carries 6 to 12 labels, and an even choice gives the first letter 9.0\% on average. Each of the 62 photos is shown with two random label orders, which gives 124 photo shelves and 124 masked shelves per model, and Qwen3.5 0.8B sees 26 of the photos. Upside-down copies of the photos and of the masked photos separate position from content for GPT-6 Luna and Qwen3.5 0.8B, the two models that favor the bottom of a photo. The photo prompt names no shopping category, since one photo may contain several categories.

\subsection{Prompts}
\label{sec:suppl-prompts}

\paragraph{Shelf image.} ``You are a shopper in a supermarket looking for \{need\}. The image shows the shelf in front of you; every product has a letter label. Which one product do you put in your basket? Answer with the letter only.'' This prompt is also used on the shelves with price tags for the shopper without a budget and on the dominance scenes.

\paragraph{Shelf image with price tags.} ``You are a shopper in a supermarket looking for \{need\}. The image shows the shelf in front of you; every product has a letter label and a price tag. Which one product do you put in your basket? Answer with the letter only.''

\paragraph{Budget shopper.} ``You are a shopper on a tight budget in a supermarket looking for \{need\}. The image shows the shelf in front of you; every product has a letter label and a price tag. Which one product do you put in your basket? Answer with the letter only.''

\paragraph{Label-free shelf image.} ``You are a shopper in a supermarket looking for \{need\}. The image shows the shelf in front of you. Which one product do you put in your basket? Answer with the product name only.'' An answer counts when its words identify the name of a single product.

\paragraph{Text list.} ``You are a shopper in a supermarket looking for \{need\}. The shelf in front of you holds these products, each with a letter label:'' followed by one line per shelf level and ``Which one product do you put in your basket? Answer with the letter only.''

\paragraph{Shelves as separate images.} ``You are a shopper in a supermarket looking for \{need\}. The images show the shelves of the unit in front of you, one image per shelf, in no particular order; every product has a letter label. Which one product do you put in your basket? Answer with the letter only.''

\paragraph{Real photo.} ``You are a shopper in a supermarket. The image shows the shelf in front of you; some products carry a letter label. Which one of the labelled products do you put in your basket? Answer with the letter only.''

\paragraph{Label reading.} ``The image shows a supermarket shelf; every product has a letter label. Which letter labels the product ``\{name\}''? Answer with the letter only.''

\paragraph{Price reading.} ``The image shows a supermarket shelf; every product has a letter label and a price tag. Which letter labels the product that costs \{price\}? Answer with the letter only.''

\paragraph{Paraphrases.} ``Here is a supermarket shelf. You need \{need\}. Each product is marked with a letter. Reply with the letter of the product you would buy and nothing else.'' and ``Imagine you are shopping for \{need\} and see this shelf. Which product would you pick? Give only its letter.''

\paragraph{Goal-directed choice.} The shelf-image prompt with a specific need in place of the category. The needs, with the product that meets them, are the diet version of a classic dark soda (zero cola) and an unsweetened bubbly water with a citrus note (lemon sparkling water); something to protect your skin from sunburn at the beach (sunscreen) and something to cover a small cut on your finger (adhesive bandages); a small pastry with a baked custard filling (egg tart) and a soft bun with a sweet azuki paste filling (red bean bun); a dairy-free frozen dessert that tastes of citrus (lemon sorbet) and a savory Italian snack you can heat in the oven (frozen pizza slice); something to power a TV remote (AA batteries) and something to clean greasy plates in the sink (dish soap); a fermented spicy Korean side dish (kimchi) and a soy-based protein for a vegetarian stir-fry (tofu); something to check whether you have a fever (digital thermometer) and something to sort a week of tablets by day (weekly pill organizer); something to check that a shelf hangs perfectly level (spirit level) and something to protect your eyes while drilling (safety glasses); something to cut back the branches of a rose bush (pruning shears) and a container to carry water to potted plants (watering can); a dairy-free milk that foams well for a latte (barista oat milk) and a mix for a Middle Eastern fried street food (falafel mix); a bittersweet citrus spread for toast (orange marmalade) and something made from tree sap to pour over pancakes (maple syrup); coffee you can drink late in the evening without losing sleep (decaf coffee) and a calming herbal tea for bedtime (chamomile tea). Each need describes its product without naming it, and no other product of the category meets it.

\section{Factor Model and Estimation}
\label{sec:suppl-model}

\paragraph{Identification.}
Under the additive model of Eq.~(\ref{eq:factor}), the three effects are identified by the design of the renderings. Within a category, the Latin square crosses products and positions, and the three label orders cross letters and positions, since reading order places A at the top left, reverse order at the bottom right, and random order at any position. The blank renderings remove the product effect by construction, the masked photos hide what each product shows, and the label-free and text renderings remove the labels or the image. Each effect is centered, which leaves the model without a free constant, and contrasts between groups of positions or letters do not depend on the centering.

\paragraph{Estimation.}
For a model that gives letter probabilities (\cref{sec:suppl-bench}), the fit maximizes $\sum_n \sum_j p_{nj} \log P_n(j)$, the expected log likelihood under the model's own letter distribution, plus the log of the Gaussian priors below. For a model that gives one answer, $p_n$ places all probability mass on the chosen letter, and shelves without a valid answer do not enter the fit. Fitting the full distribution uses the belief of the model over all 12 options from every query, whereas a single answer gives one observed choice. Gaussian priors with standard deviation 4 on the label and layout effects and 1.5 on the product effects keep the fit finite when an option is never chosen.

\paragraph{Model checking.}
Nested factor models and prior widths are scored by the log probability of held-out answers, with folds that hold out a quarter of the placements in every category (\cref{tab:suppl-modelcheck}). For the hosted models, adding label and layout effects to the product effects improves the prediction of held-out answers by 0.125 to 0.568 nats per shelf, and the full model also predicts better than label and layout effects alone. Interactions between the layout effect and the label order, or between the label effect and the shelf, add at most 0.011 nats and lower the score of the models that give one answer. The additive model of Eq.~(\ref{eq:factor}) therefore suffices. A prior of 2 or 8 on the label and layout effects changes the held-out score by at most 0.002 nats per shelf. Among the four widths of the prior on the product effects, 1.5 is within 0.015 nats of the best for every model, while a narrower prior of 0.75 lowers the held-out score of the hosted models by 0.058 to 0.078 nats.

\begin{table}[!htbp]
\centering
\small
\resizebox{\linewidth}{!}{%
\begin{tabular}{lccccc}
\toprule
Model & LFM2.5-VL 0.45B & Qwen3.5 0.8B & Qwen3.8-Flash & Claude Haiku 5.5 & GPT-6 Luna \\
\midrule
Product only & $-2.500$ & $-2.503$ & $-1.815$ & $-1.798$ & $-2.028$ \\
Label + layout & $-2.285$ & $-1.985$ & $-2.469$ & $-2.377$ & $-2.131$ \\
Label + layout + product & $-2.282$ & $-1.976$ & $-1.690$ & $\mathbf{-1.504}$ & $\mathbf{-1.460}$ \\
With layout by label order & $-2.274$ & $-1.968$ & $\mathbf{-1.682}$ & $-1.542$ & $-1.486$ \\
With label by shelf & $\mathbf{-2.273}$ & $\mathbf{-1.966}$ & $-1.689$ & $-1.557$ & $-1.531$ \\
\midrule
Prior widths 4 / 0.75 & $\mathbf{-2.282}$ & $-1.977$ & $-1.768$ & $-1.577$ & $-1.518$ \\
Prior widths 4 / 1.5 & $\mathbf{-2.282}$ & $\mathbf{-1.976}$ & $-1.690$ & $-1.504$ & $-1.460$ \\
Prior widths 4 / 3 & $-2.283$ & $-1.978$ & $-1.676$ & $-1.569$ & $-1.514$ \\
Prior widths 4 / 6 & $-2.283$ & $-1.978$ & $\mathbf{-1.675}$ & $-1.690$ & $-1.611$ \\
Prior widths 2 / 1.5 & $\mathbf{-2.282}$ & $\mathbf{-1.976}$ & $-1.690$ & $\mathbf{-1.502}$ & $\mathbf{-1.459}$ \\
Prior widths 8 / 1.5 & $\mathbf{-2.282}$ & $\mathbf{-1.976}$ & $-1.690$ & $-1.505$ & $-1.460$ \\
\bottomrule
\end{tabular}}
\caption{Mean log probability of held-out answers per shelf (higher is better; an even choice gives $-2.485$), with folds that hold out a quarter of the placements in every category. The upper block compares nested factor models with prior widths 4 (label and layout) and 1.5 (product), and the lower block compares prior widths for the full model, given as label and layout / product.}
\label{tab:suppl-modelcheck}
\end{table}

\paragraph{Reliability correction.}
For each rendering, the product effects are fitted separately on the even- and odd-numbered arrangements, and the correlation $r_h$ between the two halves, corrected to the full number of arrangements by the Spearman-Brown formula $2r_h/(1+r_h)$, gives the split-half reliability $\rho$ of the rendering. The agreement between two renderings $a$ and $b$ is then $r_{ab}/\sqrt{\rho_a \rho_b}$, an estimate of the correlation that the product effects of the two renderings would show without estimation noise. A value near one means that the two renderings rank the products the same way up to noise, and since the reliabilities are themselves estimated, a value can exceed one. A reliability below 0.3 makes the corrected value unstable, and such pairs are left out. Without this correction a rendering with fewer shelves would appear less stable only because of noisier effects.

\paragraph{Intervals.}
All intervals, given in brackets, are 95\% percentile intervals from resampling categories with replacement (photos for the real shelf photos), with the full fit repeated in every resample. Paired differences between two corrections are resampled jointly, and the intervals of these differences thereby account for the correlation between the corrections.

\subsection{Human Reference Effects}
\label{sec:suppl-human}

Both human studies that we compare with asked people to choose from shelf images, a task like the one \bench poses to a model, and both report choice shares by position. In the study of Chandon et al.~\cite{chandon2009does}, 348 shoppers viewed life-size projections of shelf photographs with an eye tracker and then made a choice. Products at the center of a display were noticed by 80\% of shoppers against 65\% at the sides, on the middle shelves by 80\% against 64\%, and on the top shelf by 74\% against 70\% on the bottom shelf, with no difference between left and right, and doubling the facings raised noticing from 60\% to 76\%. As log ratios, the five noticing effects are $+0.21$, $+0.22$, $+0.06$, $0.00$, and $+0.24$, and the attention prior is fitted to these five effects. The choice share rose from 7.7\% to 9.0\% at the center and from 7.9\% to 8.8\% on the top shelf, the latter not significant, which gives log ratios of $+0.16$ and $+0.11$. Choice rose by $+0.13$ with 4 to 12 facings. Atalay et al.~\cite{atalay2012shining} showed three-column displays on a computer screen and found 45.3\% of choices in the center column, a log ratio of $+0.50$ against the mean of the two side columns. We express the choice shares of both studies as log ratios between positions, which at choice shares near 1/12, as in both studies, closely approximate the contrasts of the position effects in Eq.~(\ref{eq:factor}). Under the conditional logit, these contrasts equal the change in the log-odds of a product against any other option and do not depend on the other options. We therefore set the human center effect between $+0.16$ and $+0.50$ and the human top-bottom effect at $+0.11$.

\subsection{Valid Answers}
\label{sec:suppl-answers}

Qwen3.8-Flash and GPT-6 Luna give a valid answer on almost every shelf (\cref{tab:suppl-answers}). The analysis of Claude Haiku 5.5 covers the rendered shelves with products, on which Claude Haiku 5.5 answers on most shelves, and leaves out the blank shelves and the real photos.

\begin{table}[!htbp]
\centering
\small
\setlength{\tabcolsep}{3pt}
\resizebox{\ifdim\width>\linewidth\linewidth\else\width\fi}{!}{%
\begin{tabular}{lccc}
\toprule
Rendering & Qwen3.8-Flash & Claude Haiku 5.5 & GPT-6 Luna \\
\midrule
Photos, reading-order labels & 100 & 93 & 100 \\
Photos, reverse labels & 100 & 85 & 100 \\
Photos, random labels & 100 & 86 & 100 \\
Perspective view & 100 & 67 & 100 \\
Upside down & 100 & 76 & 100 \\
Schematic packs & 100 & 92 & 100 \\
No labels & 76 & 93 & 84 \\
Blank packs & 92 & 0 & 100 \\
Text, top first & 100 & 98 & 100 \\
Text, bottom first & 100 & 98 & 99 \\
Real photos (SKU-110K) & 100 & 17 & 100 \\
Real photos (ShelfWise) & 100 & 7 & 100 \\
Masked photos (SKU-110K) & 100 & 9 & 100 \\
Masked photos (ShelfWise) & 96 & 7 & 100 \\
Quarter turn & 100 & 78 & 100 \\
Mirrored & 100 & 82 & 100 \\
Labels M to X & 100 & 88 & 99 \\
Price tags & 100 & 86 & 100 \\
Price tags, budget shopper & 100 & 66 & 100 \\
First paraphrase & 100 & 96 & 100 \\
Second paraphrase & 100 & 96 & 100 \\
\bottomrule
\end{tabular}}
\caption{Percentage of shelves on which a hosted model gives a valid answer (a single label letter, or one product name in the label-free view). The open models give a letter distribution on every shelf.}
\label{tab:suppl-answers}
\end{table}

\section{Evidence for Each Requirement}
\label{sec:suppl-evidence}

\begin{table}[!htbp]
\centering
\small
\setlength{\tabcolsep}{4pt}
\resizebox{\ifdim\width>\linewidth\linewidth\else\width\fi}{!}{%
\begin{tabular}{llcccc}
\toprule
Requirement & Test & Qwen3.8-Flash & Claude Haiku 5.5 & GPT-6 Luna & Reference \\
\midrule
Read the shelf (R1) & Label of a named product (\%) & 94 & 94 & 93 & 8 by chance \\
 & Label of a priced product (\%) & 100 & 99 & 100 & 8 by chance \\
\midrule
Choose what is needed (R2) & Product that meets a stated need (\%) & 96 & 95 & 90 & 8 by chance \\
\midrule
Preferences of products (R3) & Perspective camera (corrected $r$) & 0.99 & 0.99 & 1.03 & 1 \\
 & Upside down (corrected $r$) & 0.95 & 0.93 & 0.73 & 1 \\
 & Quarter turn (corrected $r$) & 0.98 & 0.90 & 0.76 & 1 \\
 & Mirrored (corrected $r$) & 0.97 & 0.85 & 1.06 & 1 \\
 & Labels M to X (corrected $r$) & 1.00 & 0.99 & 0.95 & 1 \\
 & Paraphrased prompts (corrected $r$) & 0.98 & 0.96 & 0.99 & 1 \\
\midrule
Respond to appearance and price (R4) & Photo vs.\ text list (corrected $r$) & 0.56 & 0.60 & 0.58 & below 1 \\
 & Price elasticity, no budget & ${+}0.08$ & ${+}0.05$ & ${+}0.36$ &  \\
 & Price elasticity, budget shopper & $-2.28$ & $-2.69$ & $-5.64$ & $-2.62$ for sales \\
 & Cheaper of two copies, budget (\%) & 58.7 &  &  & 50 by chance \\
\midrule
Answer effects (R5) & Letter A (log-odds) & ${+}0.84$ & ${+}1.24$ & ${+}2.17$ & 0 \\
 & Top vs.\ bottom shelf (log-odds) & ${+}0.66$ & ${+}1.18$ & ${+}0.68$ & ${+}0.11$ for people \\
 & Top of the image, four views (log-odds) & ${+}0.86$ & ${+}1.28$ & ${+}0.55$ &  \\
 & Physical top shelf, four views (log-odds) & ${+}0.10$ & $-0.03$ & ${+}0.34$ &  \\
\midrule
Removable (R5) & Ranking before / after removal & 0.56 / 0.56 & 0.44 / 0.54 & 0.44 / 0.52 & 1 \\
\bottomrule
\end{tabular}}
\caption{Tests of the five requirements and results of the three hosted models. Corrected $r$ is the reliability-corrected correlation with the product-photo front views, the ranking is the rank correlation of product utilities with the fitted product effects, and the reference column gives the value for an even choice, for perfect stability, for brand sales \cite{bijmolt2005new}, or for human shoppers \cite{chandon2009does}.}
\label{tab:suppl-evidence}
\end{table}

\Cref{tab:suppl-evidence} gives the tests of the five requirements of \cref{sec:shelfsight} and the results of the three hosted models next to a reference value from chance, perfect stability, brand sales, or people. Corrected correlations are omitted when a reliability is below 0.3, and the cheaper-copy share is given where the two copies of a product receive at least 50 choices. The ranking uses the letter probabilities of one query for Qwen3.8-Flash and twelve answers for the two models that give one answer, either counted or with the answer effects removed, with the products that no answer chose kept tied in both.

Because a reading question asks for one letter, a correct answer requires finding the named product or the stated price in the image and then the adjacent letter. The goal test names a need that exactly one of the 12 products meets. The scene changes keep the products and alter the camera, the orientation of the image, the label set, the price tags, or the prompt, while photographs, names, and prices test the response to appearance and cost. Because the label and layout effects are shared across categories, the effects can be estimated on some categories and removed from the answers in other categories. The subsections below and \cref{sec:suppl-answer-effects,sec:suppl-correction} give the details of each test.

\subsection{Reading and Goal-Directed Choice}
\label{sec:suppl-reading}

\begin{table}[!htbp]
\centering
\small
\setlength{\tabcolsep}{4pt}
\begin{tabular}{lccc}
\toprule
Model & Front & Perspective & Top / bottom shelf \\
\midrule
LFM2.5-VL 0.45B & 25 & 14 & 25 / 25 \\
Qwen3.5 0.8B & 58 & 56 & 58 / 58 \\
Qwen3.8-Flash & 94 & 92 & 97 / 92 \\
Claude Haiku 5.5 & 94 & 89 & 94 / 94 \\
GPT-6 Luna & 93 & 85 & 92 / 94 \\
\bottomrule
\end{tabular}
\caption{Label reading. Percentage of questions in which the most likely answer is the letter of the named product, for front views with product photos and for the perspective view, 72 questions each. The last column splits the front-view questions by the shelf of the named product.}
\label{tab:reading}
\end{table}

The hosted models give the correct letter for a named product on 93\% to 94\% of front views (\cref{tab:reading}). On front views the hosted models read the bottom shelf with accuracy comparable to the top shelf, 92\% to 94\% against 92\% to 97\%. Asked for the letter of the product with a given price, the hosted models answer correctly on 69 or 70 of the 70 shelves. The small open models read the labels of schematic packs, which print the product name, on 88\% and 82\% of shelves.

In the goal-directed test, the hosted models choose the product that meets the stated need on 90\% to 96\% of answered shelves, against 8\% for random guessing. The rate is 91\% or more on the bottom shelf, and Qwen3.8-Flash and Claude Haiku 5.5 choose the target on every shelf where the target carries the letter A.

\subsection{Stability of Product Preferences}
\label{sec:suppl-stability}

\begin{figure}[!htbp]
\centering
\includegraphics[width=\linewidth]{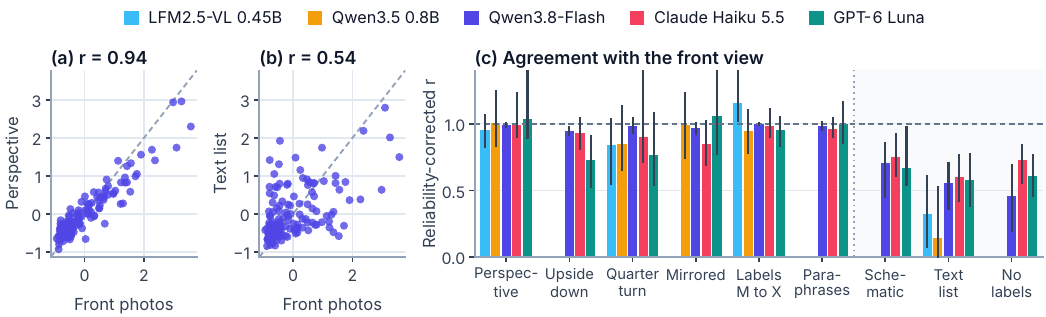}
\caption{Product preferences across renderings. (a, b) Product effects of Qwen3.8-Flash on the product-photo front views against those on the perspective view and on the text list, one point per product; the dashed line marks equality, and r is the uncorrected Pearson correlation. (c) Reliability-corrected correlation with the product-photo front views for each rendering and model; the dashed line marks one, bars are truncated at 1.4, and the shaded side holds renderings that change how the products are shown or how the model answers. Error bars are 95\% intervals.}
\label{fig:suppl-stability}
\end{figure}

\begin{table}[!htbp]
\centering
\small
\setlength{\tabcolsep}{3pt}
\resizebox{\ifdim\width>\linewidth\linewidth\else\width\fi}{!}{%
\begin{tabular}{lccccc}
\toprule
Rendering & LFM2.5-VL 0.45B & Qwen3.5 0.8B & Qwen3.8-Flash & Claude Haiku 5.5 & GPT-6 Luna \\
\midrule
Front photos & (0.70) & (0.78) & (0.97) & (0.86) & (0.84) \\
Perspective & 0.95 {\scriptsize [0.82, 1.07] (0.61)} & 1.01 {\scriptsize [0.82, 1.25] (0.65)} & 0.99 {\scriptsize [0.97, 1.01] (0.93)} & 0.99 {\scriptsize [0.90, 1.24] (0.70)} & 1.03 {\scriptsize [0.89, 1.85] (0.54)} \\
Upside down &  &  & 0.95 {\scriptsize [0.91, 0.99] (0.95)} & 0.93 {\scriptsize [0.81, 1.05] (0.76)} & 0.73 {\scriptsize [0.52, 0.91] (0.76)} \\
Quarter turn & 0.84 {\scriptsize [0.54, 1.04] (0.44)} & 0.85 {\scriptsize [0.64, 1.14] (0.47)} & 0.98 {\scriptsize [0.92, 1.05] (0.78)} & 0.90 {\scriptsize [0.71, 1.85] (0.46)} & 0.76 {\scriptsize [0.54, 1.09] (0.50)} \\
Mirrored & \scriptsize ($-0.01$) & 0.99 {\scriptsize [0.74, 1.24] (0.42)} & 0.97 {\scriptsize [0.91, 1.01] (0.90)} & 0.85 {\scriptsize [0.69, 1.03] (0.76)} & 1.06 {\scriptsize [0.76, 2.28] (0.45)} \\
Labels M to X & 1.15 {\scriptsize [1.01, 1.41] (0.63)} & 0.94 {\scriptsize [0.78, 1.11] (0.61)} & 1.00 {\scriptsize [0.99, 1.01] (0.95)} & 0.99 {\scriptsize [0.89, 1.12] (0.80)} & 0.95 {\scriptsize [0.83, 1.06] (0.77)} \\
Price tags & 1.28 {\scriptsize [1.09, 1.66] (0.52)} & 1.15 {\scriptsize [1.02, 1.35] (0.57)} & 1.01 {\scriptsize [0.99, 1.04] (0.93)} & 1.00 {\scriptsize [0.90, 1.08] (0.74)} & 0.98 {\scriptsize [0.84, 1.18] (0.71)} \\
Paraphrases &  &  & 0.98 {\scriptsize [0.95, 1.02] (0.88)} & 0.96 {\scriptsize [0.89, 1.05] (0.82)} & 0.99 {\scriptsize [0.85, 1.17] (0.59)} \\
Schematic &  &  & 0.70 {\scriptsize [0.45, 0.87] (0.88)} & 0.75 {\scriptsize [0.60, 0.93] (0.62)} & 0.67 {\scriptsize [0.54, 0.98] (0.49)} \\
Text list & 0.32 {\scriptsize [0.07, 0.62] (0.43)} & 0.14 {\scriptsize [$-0.24$, 0.53] (0.51)} & 0.56 {\scriptsize [0.35, 0.71] (0.97)} & 0.60 {\scriptsize [0.41, 0.77] (0.91)} & 0.58 {\scriptsize [0.37, 0.78] (0.74)} \\
No labels &  &  & 0.46 {\scriptsize [0.19, 0.70] (0.93)} & 0.72 {\scriptsize [0.55, 0.85] (0.94)} & 0.61 {\scriptsize [0.45, 0.77] (0.88)} \\
\bottomrule
\end{tabular}}
\caption{Correlation of product effects between the product-photo front views and each other rendering, divided by the square root of the product of the two split-half reliabilities, with 95\% intervals from resampling categories. The split-half reliability of each rendering is given in parentheses, and the first row gives that of the front views. A correlation is omitted when a reliability is below 0.3 or the rendering was not shown to the model.}
\label{tab:suppl-stability}
\end{table}

\Cref{fig:suppl-stability} and \cref{tab:suppl-stability} give the reliability-corrected correlation between the product effects of the product-photo front views and those of every other rendering, with the split-half reliability of each rendering. The renderings above the schematic row keep the products and change the scene, and 17 of the 21 corrected correlations of the hosted models exceed 0.9. Qwen3.8-Flash, whose product effects are the most reliable (0.97 on the front view), keeps the same preferences under every change of scene, at 0.95 to 1.01. The perspective view reshapes every pack, the upside-down and mirrored views reverse the text on the packs, the quarter turn rotates every pack by 90 degrees, the price tags add a new object under each product, and the labels M to X replace every letter.

The renderings below change how the products are shown or how the model answers, and agreement is lower, at 0.67 to 0.75 for schematic packs and 0.56 to 0.60 for the text list. The schematic packs keep the name and the color of a product and omit the photograph, the text list keeps only the name, and the change of answer format changes the task from picking a letter to naming a product. High agreement across scene changes and lower agreement across changes of content show that the preferences follow the product and not the position of the product. The open models also keep their preferences under the perspective view and the quarter turn (0.84 to 1.01).

\subsection{Appearance and Salience}
\label{sec:suppl-appearance}

\begin{figure}[!htbp]
\centering
\includegraphics[width=\linewidth]{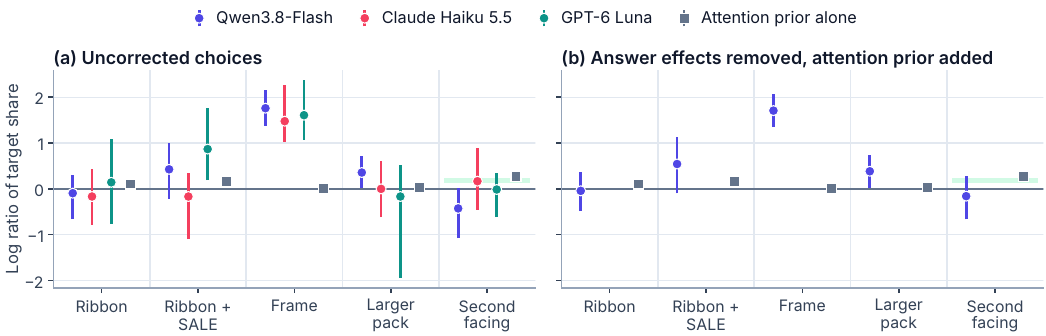}
\caption{Choice lift of a target product under each visual manipulation, as the log ratio of its mean choice share against the same scene with a plain target. (a) Uncorrected choices. (b) Choices after removing the label and layout effects and adding the attention prior, for models that give a letter distribution; squares show the change in attention share alone. The green band spans the human facings effects on choice and on noticing. Bars are 95\% intervals from resampling categories.}
\label{fig:suppl-salience}
\end{figure}

\paragraph{Photographs against names.}
The product effects of the hosted models on photo shelves agree with those on schematic packs at 0.67 to 0.75 and with those of the text list at 0.56 to 0.60 (\cref{tab:suppl-stability}). The spectral-residual saliency \cite{hou2007saliency} of a packshot explains little of this difference. This saliency correlates with the difference between a product's photo and schematic effects at $-0.02$ to $+0.05$, with all intervals including zero, and with the difference between the photo and text effects of the product at $-0.05$ to $+0.20$. The photograph thus adds content beyond the low-level contrast that saliency captures.

\paragraph{Salience.}
A yellow frame around one target product raises the log share of the target by $+1.48$ to $+1.76$ in the hosted models (\cref{fig:suppl-salience}) and thus multiplies the choice share of the target by 4.4 to 5.8, as visual marks do when prompting VLMs \cite{yang2023setofmark,shtedritski2023what}. A ribbon with a ``SALE'' bar raises the share for GPT-6 Luna by $+0.87$ [$+0.19$, $+1.77$], and a pack approximately 30\% larger raises the share for Qwen3.8-Flash by $+0.36$ [$+0.01$, $+0.71$]. For people, doubling the facings raises noticing by $+0.24$ and 4 to 12 facings raise choice by $+0.13$ \cite{chandon2009does}, and the attention prior adds this facings effect. In the salience scenes a second facing raises the log attention share of a product by approximately $+0.25$ because the product then covers more of the shelf. Combined with the attention prior, the corrected choices of Qwen3.8-Flash keep the frame effect ($+1.71$), and the prior raises the response of these choices to the second facing toward that of people.

\subsection{Price}
\label{sec:suppl-price}

\paragraph{The shopper without a budget.}
With the shopping prompt, which asks for a product of the category and does not mention prices, the hosted models choose by product while reading the price of a product on 99\% to 100\% of shelves. Whether the hosted models respond to price is thus set by the stated profile, as the budget shopper shows.

\paragraph{The budget shopper.}
On the same shelves, with only the prompt changed to a shopper on a tight budget, the elasticities are $-2.28$ [$-3.37$, $-1.78$] for Qwen3.8-Flash and $-2.69$ [$-4.20$, $-2.18$] for Claude Haiku 5.5, and GPT-6 Luna responds still more strongly. Relative to the mean share of the product over all scenes, the choice share of a product falls at each step from 0.8 to 1 to 1.25 times the list price in all three models, from 1.29 to 2.01 at the lowest price to 0.28 to 0.79 at the highest. The response thus covers the whole price range. Bijmolt et al.\ \cite{bijmolt2005new} collect 1,851 price elasticities of brand sales from published studies, with a mean of $-2.62$. As budget shoppers, Qwen3.8-Flash and Claude Haiku 5.5 are close to this mean.

\paragraph{Placebo.}
To check that the price term does not absorb effects of position or label, the factor model with a price term is also fitted on the untagged photo shelves with the same placements and labels, with the prices that the tags would have shown. This placebo fit gives elasticities between $-0.13$ and $+0.21$, all intervals including zero, which shows that the budget elasticities come from the tags.

\paragraph{Dominance.}
The dominance scenes offer the same product twice, at 0.8 and 1.25 times the list price, which makes the cheaper copy preferable on price alone. As a budget shopper, Qwen3.8-Flash chooses the cheaper copy in 58.7\% [53.8\%, 64.2\%] of the choices between the two copies on 144 shelves, above an even split, and therefore responds to price between otherwise identical packs.

\section{Answer Effects}
\label{sec:suppl-answer-effects}

This section analyzes where the label and layout effects of each model come from and how these effects change across renderings.

\subsection{Image Position and Shelf Position}
\label{sec:suppl-position}

\begin{figure}[!htbp]
\centering
\includegraphics[width=\linewidth]{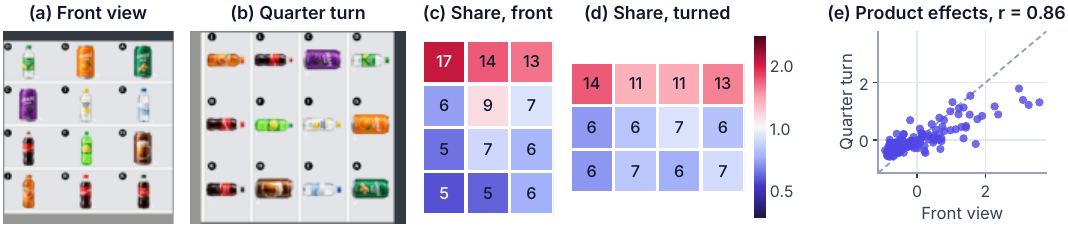}
\caption{Shelf views and the choices of Qwen3.8-Flash. (a) Front view with letter labels. (b) The same shelf turned a quarter clockwise. (c, d) Choice share, in percent, implied by the fitted layout effects at each place in the image for the two views; colors show the ratio to an even share. (e) Product effects fitted on the two views, one point per product; the dashed line marks equality, and r is the uncorrected Pearson correlation.}
\label{fig:suppl-views}
\end{figure}

\begin{table}[!htbp]
\centering
\small
\setlength{\tabcolsep}{2pt}
\resizebox{\ifdim\width>\linewidth\linewidth\else\width\fi}{!}{%
\begin{tabular}{lcccc}
\toprule
 & \multicolumn{2}{c}{Four views, one model} & \multicolumn{2}{c}{Separate shelf images} \\
\cmidrule(lr){2-3}\cmidrule(lr){4-5}
Model & Image top & Top shelf & First image & Top shelf \\
\midrule
LFM2.5-VL 0.45B & $-0.05$ {\scriptsize [$-0.16$, ${+}0.04$]} & $-0.12$ {\scriptsize [$-0.17$, $-0.06$]} &  &  \\
Qwen3.5 0.8B & $-0.36$ {\scriptsize [$-0.46$, $-0.25$]} & $-0.15$ {\scriptsize [$-0.29$, $-0.05$]} &  &  \\
Qwen3.8-Flash & ${+}0.86$ {\scriptsize [${+}0.70$, ${+}1.10$]} & ${+}0.10$ {\scriptsize [$-0.03$, ${+}0.20$]} & ${+}0.41$ {\scriptsize [${+}0.10$, ${+}0.73$]} & ${+}0.16$ {\scriptsize [$-0.07$, ${+}0.30$]} \\
Claude Haiku 5.5 & ${+}1.28$ {\scriptsize [${+}1.14$, ${+}1.68$]} & $-0.03$ {\scriptsize [$-0.40$, ${+}0.37$]} & ${+}1.43$ {\scriptsize [${+}1.11$, ${+}2.36$]} & ${+}0.01$ {\scriptsize [$-0.88$, ${+}0.96$]} \\
GPT-6 Luna & ${+}0.55$ {\scriptsize [${+}0.13$, ${+}1.05$]} & ${+}0.34$ {\scriptsize [${+}0.11$, ${+}0.66$]} & ${+}0.13$ {\scriptsize [$-0.76$, ${+}0.81$]} & $-0.08$ {\scriptsize [$-0.51$, ${+}0.19$]} \\
\bottomrule
\end{tabular}}
\caption{Position effects of the image and of the shelf, in log-odds with 95\% intervals from resampling categories. The first two columns come from one factor model over four views, and the last two from shelves shown as four separate images. \emph{Image top} spans the image from the bottom to the top edge, \emph{Top shelf} compares the physical top and bottom shelves, and \emph{First image} compares the shelf shown first with the other three. Empty cells have no data.}
\label{tab:views}
\end{table}

\paragraph{Image or shelf.}
The upside-down, mirrored, and quarter-turned views separate image position from shelf position (\cref{fig:suppl-views}), and one factor model over the four views with random labels estimates the contribution of each (\cref{tab:views}). The four views place the physical top shelf at the top, the bottom, and the side of the image. Image position and shelf position are therefore no longer tied, and the joint model can attribute the advantage to either. The joint model has a position effect for each of the 12 physical shelf positions and two image terms, linear in the height of a product in the image and in the horizontal place of the product. For Qwen3.8-Flash and Claude Haiku 5.5 the top of the image gains $+0.86$ and $+1.28$ log-odds over the bottom of the image while the physical top shelf gains $+0.10$ and $-0.03$ over the bottom shelf. Almost all of the advantage therefore follows the image. GPT-6 Luna divides the vertical advantage between the image ($+0.55$) and the shelf ($+0.34$), and both Claude Haiku 5.5 and GPT-6 Luna also favor the left of the image ($+0.71$ and $+0.43$). Among the open models, LFM2.5-VL 0.45B keeps a small preference for the physical bottom shelf, and Qwen3.5 0.8B favors the bottom and the right of the image.

\begin{table}[!htbp]
\centering
\small
\setlength{\tabcolsep}{4pt}
\resizebox{\ifdim\width>\linewidth\linewidth\else\width\fi}{!}{%
\begin{tabular}{lccccc}
\toprule
Model and crop & Shelves & First image & Last image & Top shelf & Letter A \\
\midrule
Qwen3.8-Flash, with shelf edges & 139 & ${+}0.31$ {\scriptsize [${+}0.06$, ${+}0.58$]} &  & ${+}0.30$ {\scriptsize [${+}0.05$, ${+}0.53$]} & ${+}1.26$ {\scriptsize [${+}1.13$, ${+}1.51$]} \\
Qwen3.8-Flash, without shelf edges & 138 & ${+}0.41$ {\scriptsize [${+}0.10$, ${+}0.73$]} & ${+}0.17$ {\scriptsize [$-0.03$, ${+}0.35$]} & ${+}0.16$ {\scriptsize [$-0.07$, ${+}0.30$]} & ${+}1.34$ {\scriptsize [${+}1.20$, ${+}1.60$]} \\
Claude Haiku 5.5, with shelf edges & 90 & ${+}1.43$ {\scriptsize [${+}1.11$, ${+}2.36$]} &  & ${+}0.01$ {\scriptsize [$-0.88$, ${+}0.96$]} & ${+}1.72$ {\scriptsize [${+}1.52$, ${+}2.63$]} \\
GPT-6 Luna, without shelf edges & 144 & ${+}0.13$ {\scriptsize [$-0.76$, ${+}0.81$]} & ${+}0.40$ {\scriptsize [$-0.21$, ${+}0.98$]} & $-0.08$ {\scriptsize [$-0.51$, ${+}0.19$]} & ${+}2.00$ {\scriptsize [${+}1.80$, ${+}2.79$]} \\
\bottomrule
\end{tabular}}
\caption{Shelves shown as four separate images in a random order, in log-odds with 95\% intervals from resampling categories. \emph{Shelves} counts the answered shelves, \emph{First image} and \emph{Last image} give the advantages of the shelf shown first or last over the other three, \emph{Top shelf} gives the difference between the physical top and bottom shelf, and \emph{Letter A} gives the label effect of A against the average letter.}
\label{tab:suppl-tiles}
\end{table}

\paragraph{Shelves as separate images.}
When the shelves are shown as four images in a random order, the image order and the physical shelf are crossed (\cref{tab:suppl-tiles}). Claude Haiku 5.5, evaluated on the crops with shelf edges, gives the shelf shown first an advantage of $+1.43$ over the other three and the physical top shelf none ($+0.01$), even though the dark top edge of the unit marks the top shelf. Qwen3.8-Flash favors the first image with both crops ($+0.31$ and $+0.41$), and without shelf edges, where no edge marks the top shelf, the top-shelf advantage of Qwen3.8-Flash decreases to $+0.16$. GPT-6 Luna favors neither the first image nor the top shelf on the crops without shelf edges. The letter A keeps an advantage in every case ($+1.26$ to $+2.00$).

\paragraph{Text lists.}
In a text list the order of the lines takes the place of image position. Claude Haiku 5.5 and GPT-6 Luna favor the top shelf when listed first ($+1.03$, $+1.28$) and not when listed last ($+0.04$, $-0.35$). The advantage of both models is therefore attached to the first line of the list. Qwen3.8-Flash favors the top shelf by $+0.32$ in both orders, a small preference for products listed on the top shelf.

\paragraph{Label order.}
Within reading order or reverse order alone, every letter stands at one fixed position. The random order separates letters from positions, and so does a joint fit of all three orders with one set of label effects. In that pooled fit, a separate layout effect for each label order adds at most 0.008 nats per shelf to the held-out score (\cref{tab:suppl-modelcheck}). The position effect is therefore nearly the same whichever order the letters follow, and a label effect estimated once applies to every order.

\paragraph{Real shelf photos.}
Because a product never moves on a real photo, the position and the content of the product are confounded, and the upside-down copies separate the two. GPT-6 Luna favors the lower part of an upright photo, with or without the products ($-1.54$ and $-0.98$). When the same photos are inverted, GPT-6 Luna favors the upper physical part of the photos, which then lies at the bottom of the image ($+2.02$ and $+1.10$). Qwen3.5 0.8B shows the same pattern, and in both models the advantage thus follows the bottom of the image and not the products on the lower shelves. Qwen3.8-Flash shows little vertical preference on photos, and LFM2.5-VL 0.45B favors the lower part of the photos. Every model favors the letter A on photos, GPT-6 Luna most strongly ($+3.22$).

\subsection{Label Effects}
\label{sec:suppl-letters}

\begin{figure}[!htbp]
\centering
\includegraphics[width=\linewidth]{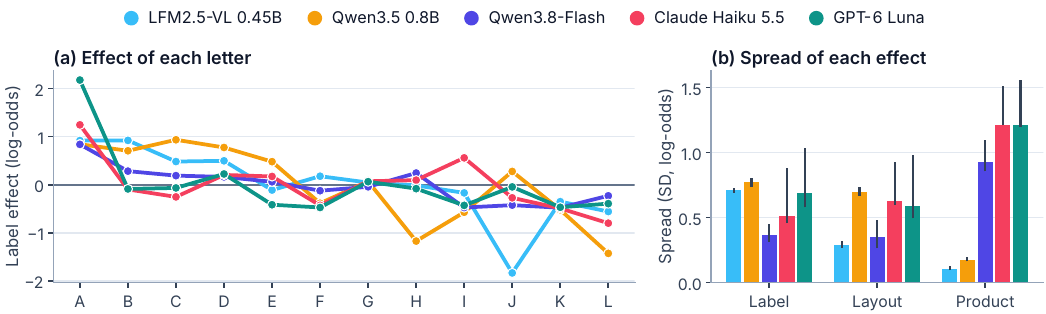}
\caption{Factor model fitted on the product-photo renderings with labels in reading, reverse, and random order. (a) Label effect of each letter, centered over the 12 letters. (b) Standard deviation of the label effects over letters, of the layout effects over positions, and of the product effects over the products of a category; bars are 95\% intervals from resampling categories.}
\label{fig:suppl-labels}
\end{figure}

All five models favor the letter A (\cref{fig:suppl-labels}), and the open models favor the first four letters approximately equally. Beyond A the letter effects of the hosted models are small and irregular. For the hosted models the product effects vary more than the label and layout effects together, while for the open models label and layout account for almost all of the variation.

With the labels M to X, the first letter M gains $+0.05$ to $+0.69$ in the hosted models, against $+0.77$ to $+2.39$ for A on the same shelves with A to L. The strong effect is therefore specific to the letter A and not a property of the first label of a set. The spread of the label effects stays at 0.32 to 0.55 against 0.35 to 0.85 with A to L.

\subsection{Effects by Rendering}
\label{sec:suppl-effects}

\begin{figure}[!htbp]
\centering
\includegraphics[width=\linewidth]{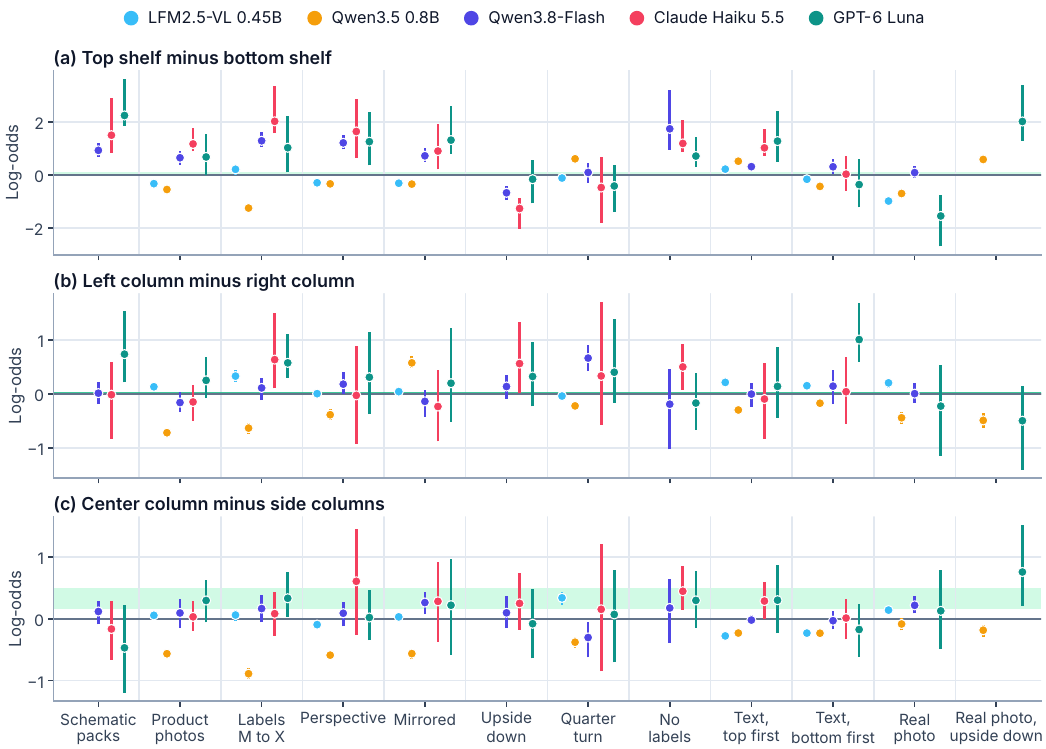}
\caption{Layout effects of each model under each rendering, as differences of the fitted position effects between groups of positions (log-odds). (a) Top shelf minus bottom shelf. (b) Left column minus right column. (c) Center column minus the two side columns. Positions are physical shelf positions in every rendering. Bars are 95\% intervals from resampling categories (photos for the real renderings). The green band spans the human choice effects of Chandon et al.~\cite{chandon2009does} and Atalay et al.~\cite{atalay2012shining}, and a green line marks a contrast where people show no difference.}
\label{fig:suppl-layout}
\end{figure}

Positions are physical shelf positions in every rendering. In the upside-down view the top shelf therefore appears at the bottom of the image, in the quarter turn at the right of the image, in the mirrored view the left column appears on the right, and in the bottom-first text the bottom shelf is listed first. The letter A is favored under every rendering that shows the letters A to L, including the text lists and the real photos (\cref{fig:suppl-layout,tab:suppl-effects,tab:suppl-effects-hosted}). The top-shelf advantage of the hosted models appears on every upright rendered shelf, with product photos, schematic packs, price tags, the labels M to X, the perspective view, and no labels ($+0.66$ to $+2.25$), and reverses or vanishes whenever the image is inverted or turned. The center column gains $+0.09$ and $+0.03$ for Qwen3.8-Flash and Claude Haiku 5.5 and $+0.30$ for GPT-6 Luna, against $+0.16$ to $+0.50$ for people, a gap that the attention prior narrows for the models that give letter probabilities (\cref{sec:suppl-attention}).

\begin{table}[!htbp]
\centering
\small
\begin{tabular}{lrrrrr}
\toprule
Rendering & Shelves & First letter & Top vs bottom & Left vs right & Center vs sides \\
\midrule
\multicolumn{6}{l}{\textit{LFM2.5-VL 0.45B}} \\
\quad Blank packs & 576 & ${+}0.52$ & $-0.40$ & ${+}0.66$ & ${+}0.24$ \\
\quad Product photos & 864 & ${+}0.92$ & $-0.32$ & ${+}0.13$ & ${+}0.06$ \\
\quad Perspective view & 288 & ${+}0.91$ & $-0.29$ & $0.00$ & $-0.09$ \\
\quad Mirrored & 72 & ${+}0.84$ & $-0.30$ & ${+}0.04$ & ${+}0.03$ \\
\quad Quarter turn & 72 & ${+}0.74$ & $-0.11$ & $-0.04$ & ${+}0.34$ \\
\quad Labels M to X & 144 & $-0.04$ & ${+}0.22$ & ${+}0.33$ & ${+}0.06$ \\
\quad Text, top first & 288 & ${+}0.62$ & ${+}0.23$ & ${+}0.21$ & $-0.28$ \\
\quad Text, bottom first & 288 & ${+}0.48$ & $-0.15$ & ${+}0.15$ & $-0.23$ \\
\quad Real photo, masked & 124 & ${+}0.87$ & $-0.95$ & ${+}0.22$ & ${+}0.16$ \\
\quad Real photo & 124 & ${+}0.88$ & $-0.98$ & ${+}0.20$ & ${+}0.14$ \\
\multicolumn{6}{l}{\textit{Qwen3.5 0.8B}} \\
\quad Blank packs & 288 & ${+}1.00$ & $-0.54$ & $-0.09$ & $-0.87$ \\
\quad Product photos & 432 & ${+}0.84$ & $-0.54$ & $-0.72$ & $-0.56$ \\
\quad Perspective view & 144 & ${+}1.16$ & $-0.33$ & $-0.39$ & $-0.59$ \\
\quad Mirrored & 72 & ${+}0.87$ & $-0.34$ & ${+}0.57$ & $-0.56$ \\
\quad Quarter turn & 72 & ${+}1.19$ & ${+}0.62$ & $-0.22$ & $-0.38$ \\
\quad Labels M to X & 144 & ${+}0.65$ & $-1.24$ & $-0.64$ & $-0.89$ \\
\quad Text, top first & 144 & ${+}0.87$ & ${+}0.53$ & $-0.30$ & $-0.23$ \\
\quad Text, bottom first & 144 & ${+}0.84$ & $-0.42$ & $-0.17$ & $-0.23$ \\
\quad Real photo, masked & 52 & ${+}1.20$ & $-0.70$ & $-0.47$ & $-0.14$ \\
\quad Real photo & 52 & ${+}1.17$ & $-0.69$ & $-0.44$ & $-0.08$ \\
\bottomrule
\end{tabular}
\caption{Label and layout effects (log-odds) of the open models for every rendering, from the factor model fitted separately on each rendering. \emph{Shelves} counts the answered decisions, and \emph{First letter} gives the label effect of A against the average letter, of M with the labels M to X, and, in the no-label rendering, a placebo: the effect of the letter A that is assigned to a product but not shown. Positions are physical shelf positions.}
\label{tab:suppl-effects}
\end{table}

\begin{table}[!htbp]
\centering
\small
\begin{tabular}{lrrrrr}
\toprule
Rendering & Shelves & First letter & Top vs bottom & Left vs right & Center vs sides \\
\midrule
\multicolumn{6}{l}{\textit{Qwen3.8-Flash}} \\
\quad Blank packs & 133 & ${+}2.58$ & ${+}0.29$ & $-0.21$ & ${+}0.14$ \\
\quad Schematic packs & 144 & ${+}1.14$ & ${+}0.93$ & ${+}0.01$ & ${+}0.12$ \\
\quad Product photos & 432 & ${+}0.84$ & ${+}0.66$ & $-0.16$ & ${+}0.09$ \\
\quad Perspective view & 144 & ${+}1.07$ & ${+}1.22$ & ${+}0.18$ & ${+}0.09$ \\
\quad Mirrored & 72 & ${+}0.78$ & ${+}0.73$ & $-0.14$ & ${+}0.26$ \\
\quad Upside down & 144 & ${+}0.79$ & $-0.66$ & ${+}0.14$ & ${+}0.10$ \\
\quad Quarter turn & 72 & ${+}0.79$ & ${+}0.10$ & ${+}0.66$ & $-0.30$ \\
\quad Labels M to X & 144 & ${+}0.23$ & ${+}1.30$ & ${+}0.11$ & ${+}0.17$ \\
\quad No labels & 109 & ${+}0.32$ & ${+}1.75$ & $-0.19$ & ${+}0.18$ \\
\quad Text, top first & 144 & ${+}0.95$ & ${+}0.32$ & $0.00$ & $-0.02$ \\
\quad Text, bottom first & 144 & ${+}1.00$ & ${+}0.32$ & ${+}0.14$ & $-0.03$ \\
\quad Real photo, masked & 123 & ${+}2.09$ & ${+}0.24$ & $0.00$ & ${+}0.30$ \\
\quad Real photo & 124 & ${+}1.76$ & ${+}0.10$ & $0.00$ & ${+}0.22$ \\
\multicolumn{6}{l}{\textit{Claude Haiku 5.5}} \\
\quad Schematic packs & 132 & ${+}2.14$ & ${+}1.51$ & $-0.02$ & $-0.16$ \\
\quad Product photos & 380 & ${+}1.24$ & ${+}1.18$ & $-0.15$ & ${+}0.03$ \\
\quad Perspective view & 97 & ${+}2.10$ & ${+}1.65$ & $-0.03$ & ${+}0.61$ \\
\quad Mirrored & 59 & ${+}1.76$ & ${+}0.91$ & $-0.23$ & ${+}0.28$ \\
\quad Upside down & 109 & ${+}1.85$ & $-1.26$ & ${+}0.56$ & ${+}0.25$ \\
\quad Quarter turn & 56 & ${+}1.25$ & $-0.47$ & ${+}0.33$ & ${+}0.15$ \\
\quad Labels M to X & 126 & ${+}0.05$ & ${+}2.03$ & ${+}0.63$ & ${+}0.08$ \\
\quad No labels & 134 & $-0.43$ & ${+}1.20$ & ${+}0.50$ & ${+}0.45$ \\
\quad Text, top first & 141 & ${+}0.77$ & ${+}1.03$ & $-0.10$ & ${+}0.29$ \\
\quad Text, bottom first & 141 & ${+}1.20$ & ${+}0.04$ & ${+}0.04$ & ${+}0.01$ \\
\multicolumn{6}{l}{\textit{GPT-6 Luna}} \\
\quad Blank packs & 144 & ${+}4.86$ & ${+}0.18$ & ${+}0.15$ & ${+}0.30$ \\
\quad Schematic packs & 144 & ${+}2.80$ & ${+}2.25$ & ${+}0.74$ & $-0.47$ \\
\quad Product photos & 432 & ${+}2.17$ & ${+}0.68$ & ${+}0.25$ & ${+}0.30$ \\
\quad Perspective view & 144 & ${+}2.34$ & ${+}1.26$ & ${+}0.31$ & ${+}0.02$ \\
\quad Mirrored & 72 & ${+}2.10$ & ${+}1.32$ & ${+}0.20$ & ${+}0.22$ \\
\quad Upside down & 144 & ${+}1.37$ & $-0.15$ & ${+}0.32$ & $-0.08$ \\
\quad Quarter turn & 72 & ${+}1.88$ & $-0.41$ & ${+}0.40$ & ${+}0.07$ \\
\quad Labels M to X & 143 & ${+}0.69$ & ${+}1.03$ & ${+}0.57$ & ${+}0.33$ \\
\quad No labels & 121 & ${+}0.16$ & ${+}0.72$ & $-0.17$ & ${+}0.30$ \\
\quad Text, top first & 144 & ${+}2.71$ & ${+}1.28$ & ${+}0.14$ & ${+}0.30$ \\
\quad Text, bottom first & 143 & ${+}2.65$ & $-0.35$ & ${+}1.01$ & $-0.17$ \\
\quad Real photo, masked & 124 & ${+}2.63$ & $-0.98$ & $-0.46$ & ${+}0.54$ \\
\quad Real photo & 124 & ${+}3.22$ & $-1.54$ & $-0.23$ & ${+}0.13$ \\
\bottomrule
\end{tabular}
\caption{Label and layout effects (log-odds) of the hosted models for every rendering, from the factor model fitted separately on each rendering. \emph{Shelves} counts the answered decisions, and \emph{First letter} gives the label effect of A against the average letter, of M with the labels M to X, and, in the no-label rendering, a placebo: the effect of the letter A that is assigned to a product but not shown. Positions are physical shelf positions.}
\label{tab:suppl-effects-hosted}
\end{table}

\section{Removing the Answer Effects}
\label{sec:suppl-correction}

\paragraph{References.}
We judge a correction by how well the product utilities from one answer rank the products of a held-out category. The first reference consists of the product effects of the factor model fitted on the other 35 renderings of the same category, which uses every view and label order of the category. The second reference, the average reference, uses no factor model. The average reference is each product's mean log probability over the perspective, price-tag, and M to X renderings, in which the Latin square balances every product over all positions. The average reference excludes the renderings with labels in reading and reverse order, which are scored against this reference. Scoring against one reference built with the factor model and one built without the factor model makes the gain of removal independent of the factor model.

\paragraph{Baselines.}
Label-only debiasing subtracts $\hat\alpha$ alone, as in the prior estimation over option permutations of PriDe \cite{zheng2024large}. Batch calibration \cite{zhou2024batch} subtracts the log of the mean letter distribution over all shelves of the test category. Content-free calibration \cite{zhao2021calibrate} subtracts the log letter distribution on the blank shelf with the same labels and positions, which costs one extra query per scene. All corrections are scored on the same shelves. The three baselines represent the removal of label effects alone, calibration on the test data, and calibration with an extra query.

\paragraph{Results.}
For the two open models, removing label and layout effects estimated on the other categories raises the rank correlation of a single answer from 0.08 to 0.24 for LFM2.5-VL 0.45B and from 0.12 to 0.27 for Qwen3.5 0.8B against the fitted reference, and from 0.04 to 0.10 and from 0.08 to 0.20 against the average reference. Removal outperforms label-only debiasing, batch calibration, and content-free calibration for both open models against both references, with paired differences of $+0.01$ to $+0.13$ whose intervals all exclude zero. Averaged over $k$ renderings of a shelf, the corrected answers of LFM2.5-VL 0.45B reach 0.56 at 12 renderings against 0.27 uncorrected. Content-free calibration performs worse because a blank shelf carries different label preferences, with the letter A at $+2.58$ on blank packs against $+0.84$ on photos for Qwen3.8-Flash. For Qwen3.8-Flash, whose single answer already ranks products at 0.56, removal keeps this value while label-only debiasing, batch calibration, and content-free calibration lower the ranking to 0.51 to 0.54 (paired differences $+0.02$ to $+0.05$ in favor of removal against the fitted reference, intervals excluding zero). Removal also raises the share of shelves on which the most preferred product agrees with the fitted reference from 58\% to 66\%.

\begin{table}[!htbp]
\centering
\small
\setlength{\tabcolsep}{4pt}
\resizebox{\ifdim\width>\linewidth\linewidth\else\width\fi}{!}{%
\begin{tabular}{lcccc}
\toprule
Answers per shelf & 2 & 4 & 6 & 12 \\
\midrule
Claude Haiku 5.5, counting & \textbf{0.37 {\scriptsize [0.27, 0.44]}} & 0.47 {\scriptsize [0.38, 0.54]} & 0.52 {\scriptsize [0.43, 0.59]} & 0.44 {\scriptsize [0.35, 0.54]} \\
Claude Haiku 5.5, removal, unchosen products ranked by answer effects & 0.24 {\scriptsize [0.17, 0.31]} & 0.33 {\scriptsize [0.25, 0.41]} & 0.40 {\scriptsize [0.30, 0.47]} & 0.40 {\scriptsize [0.31, 0.48]} \\
Claude Haiku 5.5, removal, unchosen products tied & \textbf{0.37 {\scriptsize [0.27, 0.45]}} & \textbf{0.49 {\scriptsize [0.39, 0.56]}} & \textbf{0.54 {\scriptsize [0.45, 0.61]}} & \textbf{0.54 {\scriptsize [0.43, 0.63]}} \\
GPT-6 Luna, counting & \textbf{0.31 {\scriptsize [0.22, 0.40]}} & 0.40 {\scriptsize [0.31, 0.49]} & 0.42 {\scriptsize [0.29, 0.54]} & 0.44 {\scriptsize [0.33, 0.56]} \\
GPT-6 Luna, removal, unchosen products ranked by answer effects & 0.20 {\scriptsize [0.14, 0.26]} & 0.32 {\scriptsize [0.23, 0.39]} & 0.36 {\scriptsize [0.26, 0.45]} & 0.41 {\scriptsize [0.32, 0.50]} \\
GPT-6 Luna, removal, unchosen products tied & \textbf{0.31 {\scriptsize [0.22, 0.41]}} & \textbf{0.42 {\scriptsize [0.33, 0.51]}} & \textbf{0.45 {\scriptsize [0.32, 0.57]}} & \textbf{0.52 {\scriptsize [0.42, 0.62]}} \\
\bottomrule
\end{tabular}}
\caption{Models that give one answer. Rank correlation between product utilities from $k$ answers on random renderings of one held-out shelf and the product effects fitted on the remaining renderings. Utilities come from counting the answers or from removing the label and layout effects estimated on other categories, with the products that no answer chose either ranked by the answer effects of these products or left tied, as in counting. Intervals resample categories.}
\label{tab:suppl-pooled}
\end{table}

\paragraph{Models that give one answer.}
For Claude Haiku 5.5 and GPT-6 Luna, \cref{tab:suppl-pooled} compares counting the answers on $k$ renderings of a held-out shelf with fitting product utilities to the same answers after removing the label and layout effects estimated on the other categories. Counting leaves the products that no answer chose tied, and the comparison keeps these products tied after removal as well, which gives both methods the same information. Any gain of removal over counting then comes from the answer effects estimated on other categories. Removal matches or exceeds counting at every $k$, and the advantage of removal grows to 0.54 against 0.44 for Claude Haiku 5.5 and 0.52 against 0.44 for GPT-6 Luna at 12 answers.

\paragraph{Real shelf photos.}
On real photos, the effects estimated on masked copies of the other photos bring the label preference close to an even choice. On masked photos Qwen3.8-Flash assigns 45.3\% of the choices to the first letter, 31.5\% after removing the effects estimated on rendered shelves, and 13.9\% after removing those estimated on masked photos, against 9.0\% for an even choice. On unmasked photos the first letter receives 37.3\%, 24.1\%, and 10.0\%. Masked copies share the layout and the labels of the photos to be corrected, which is why masked copies calibrate real photos better than rendered shelves do.

\subsection{Attention Priors}
\label{sec:suppl-attention}

\begin{figure}[!htbp]
\centering
\includegraphics[width=\linewidth]{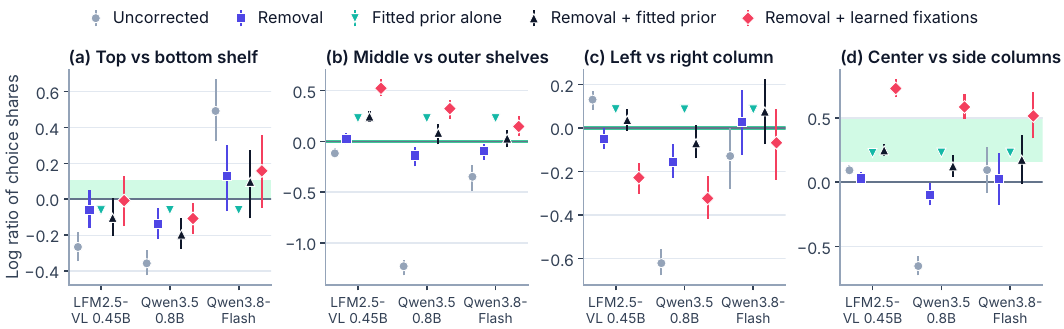}
\caption{Position effects of the synthetic shoppers on held-out categories, as log ratios of mean choice shares between groups of positions. Uncorrected answers, answers with the label and layout effects removed, the attention prior fitted on human noticing data alone, and corrected answers combined with that prior or with a fixation model learned from human eye movements on natural images. The green band spans the human choice effects of Chandon et al.~\cite{chandon2009does} and Atalay et al.~\cite{atalay2012shining}, and a green line marks a contrast where people show no difference. Bars are 95\% intervals from resampling categories.}
\label{fig:human}
\end{figure}

\begin{table}[!htbp]
\centering
\small
\setlength{\tabcolsep}{4pt}
\resizebox{\linewidth}{!}{%
\begin{tabular}{lcccccc}
\toprule
 & Center vs sides & Middle vs outer & Top vs bottom & Left vs right & Doubled facings & Center column \\
\midrule
Human noticing \cite{chandon2009does} & ${+}0.21$ & ${+}0.22$ & ${+}0.06$ & $0.00$ & ${+}0.24$ &  \\
Human choice \cite{chandon2009does,atalay2012shining} & ${+}0.16$ & n.s. & ${+}0.11$ &  & ${+}0.13$$^*$ & $45\%$ chosen \\
\midrule
Saliency only & $-0.09$ & $-0.07$ & $-0.04$ & ${+}0.10$ & ${+}0.33$ & $31\%$ \\
Fitted mixture ($w=0.28$) & ${+}0.20$ & ${+}0.23$ & $-0.05$ & ${+}0.08$ & ${+}0.34$ & $38\%$ \\
Central bias only & ${+}0.91$ & ${+}1.13$ & $-0.14$ & ${+}0.01$ & ${+}0.33$ & $55\%$ \\
Learned fixation model & ${+}0.70$ & ${+}0.52$ & $-0.01$ & $-0.26$ & ${+}0.27$ & $50\%$ \\
\bottomrule
\end{tabular}}
\caption{Position and facings effects of the attention priors on the rendered front views, as log ratios of attention shares, next to the human effects. The last column is the share of the center column. $^*$From 4 to 12 facings. The \emph{Doubled facings} column compares a product shown twice side by side with the same product shown once. n.s.: not significant \cite{chandon2009does}.}
\label{tab:suppl-attention}
\end{table}

Removing the answer effects takes out the fitted position effect of a choice, and Eq.~(\ref{eq:recompose}) multiplies the corrected choice by an attention prior fitted to human noticing effects, while the product preferences still come from the VLM. \Cref{fig:human} shows the position effects on held-out categories before and after the correction and with each prior, and \cref{tab:suppl-attention} compares the priors on the rendered front views.

Saliency alone yields a negligible center or middle-shelf effect, and a central bias alone gives effects several times larger than those of people. We fit the single mixing weight on the five human noticing effects and test the resulting prior on the human choice effects. The fitted mixture ($w=0.28$) matches the center and middle-shelf effects closely, with $+0.20$ against $+0.21$ for people and $+0.23$ against $+0.22$, and matches the top-bottom, left-right, and facings effects within about 0.1. With this prior, the corrected choices of the three models that give letter probabilities favor the center column by $+0.12$ to $+0.25$ (people $+0.16$ to $+0.50$). Combined with the prior, the choices of Qwen3.8-Flash rank products as well as the corrected choices alone (0.56), while the prior alone carries no product preference (0.02). The prior thus adds position effects and leaves the ranking unchanged. A fixation model learned from human eye movements on natural images \cite{droste2020unified,judd2009learning}, UNISAL with the released weights and nothing fitted to shelves, can serve as the prior as well. Combined with the corrected choices of Qwen3.8-Flash, this fixation model places 45.7\% of the choices in the center column, as the shoppers of Atalay et al.\ do (45.3\%) \cite{atalay2012shining}. The fixation model also reproduces the human facings effect on noticing ($+0.27$ against $+0.24$).

\end{document}